\documentclass[10pt,twocolumn,letterpaper]{article}

\usepackage{cvpr}
\usepackage{times}
\usepackage{epsfig}
\usepackage{graphicx}
\usepackage{amsmath}
\usepackage{amssymb,bm}

\usepackage{floatrow}
\usepackage{multirow}
\usepackage{algorithm}
\usepackage{color}
\usepackage{algpseudocode}
\usepackage{amsthm}
 
\theoremstyle{definition}
\newtheorem{definition}{Definition}[section]

\usepackage[caption=false]{subfig}

\usepackage[pagebackref=true,breaklinks=true,letterpaper=true,colorlinks,bookmarks=false]{hyperref}
\usepackage{array}

\cvprfinalcopy 

\def\cvprPaperID{****} 

\newcommand{\matr}[1]{#1}

\DeclareMathOperator*{\argmin}{arg\,min}
\newcolumntype{?}{!{\vrule width 2pt}}
\makeatletter
\def\Cline#1#2{\@Cline#1#2\@nil}
\def\@Cline#1-#2#3\@nil{%
  \omit
  \@multicnt#1%
  \advance\@multispan\m@ne
  \ifnum\@multicnt=\@ne\@firstofone{&\omit}\fi
  \@multicnt#2%
  \advance\@multicnt-#1%
  \advance\@multispan\@ne
  \leaders\hrule\@height#3\hfill
  \cr}
\makeatother
\begin{document}

\newcommand\error[1]{\textbf{\textcolor{red}{#1}}}
\newcommand\notice[1]{\textbf{\textcolor{green}{#1}}}
\title{Answering Image Riddles using Vision and Reasoning through \\Probabilistic Soft Logic}

\author{Somak Aditya
\qquad
Yezhou Yang
\qquad
Chitta Baral\\
Arizona State University, Tempe, AZ\\
{\tt\small \{saditya1,yz.yang,chitta\}@asu.edu}
\and
Yiannis Aloimonos\\
University of Maryland, College Park\\
{\tt\small yiannis@cs.umd.edu}
}

\maketitle

\begin{abstract}

In this work, we explore a genre of puzzles (``image riddles'') which involves a set of images and a question. 
Answering these puzzles require both capabilities involving visual detection (including object, activity recognition) and, knowledge-based or commonsense reasoning. 
We compile a dataset of 
over 3k riddles where each riddle consists of $4$ images and a groundtruth answer. The annotations are validated using crowd-sourced evaluation. We also define an automatic evaluation metric to track future progress. Our task bears similarity with the commonly known IQ tasks such as analogy solving, sequence filling that are often used to test intelligence. 


We develop a Probabilistic Reasoning-based approach that utilizes probabilistic commonsense knowledge 
to answer these riddles with a reasonable accuracy.    We demonstrate the results of our approach using both automatic and human evaluations. Our approach achieves some promising results for these riddles and provides a strong baseline for future attempts. 
We make the entire dataset and related materials publicly available to the community in ImageRiddle Website (\href{http://bit.ly/22f9Ala}{http://bit.ly/22f9Ala}).
\end{abstract}

\vspace*{-0.1in}
\section{Introduction}
\vspace*{-0.05in} 
\begin{figure}[!htpb]
     \centering
     \includegraphics[width=\textwidth]{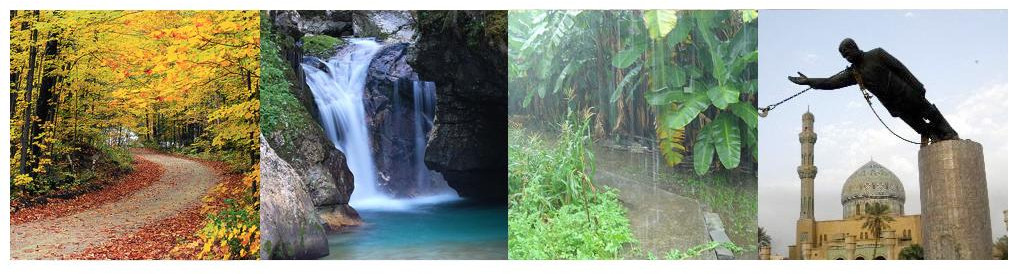}
     \caption{An Image Riddle Example. Question: ``What word connects these images?'' . } 
     \label{fig:fall}
\vspace*{-0.08in}     
\end{figure}
\vspace*{-0.08in}

A key component of computer vision is understanding of images and it comes up in various tasks such as image captioning and visual question answering (VQA). In this paper, we propose a new task of ``image riddles'' which requires deeper and conceptual understanding of images. In this task a set of images are provided and one needs to find a concept (described in words) that is invoked by all the images in that set.  Often the common concept is not something that even a human can observe in her first glance but can come up with after some thought about the images. Hence the word ``riddle'' in the phrase ``image riddles''. Figure \ref{fig:fall} shows an example of an image riddle. The images individually connect to multiple concepts such as: \textit{outdoors, nature, trees, road, forest, rainfall, waterfall, statue, rope, mosque} etc. On further thought, the common concept that emerges for this example is ``fall''. Here, the first image represents the fall season (\textit{concept}). There is a ``waterfall'' (\textit{region}) in the second image. In the third image, it shows ``rainfall'' (\textit{concept}) and the fourth image depicts that a statue is ``fall''ing (\textit{action/event}). The word ``Fall'' is invoked by all the images as it shows logical connections to objects, regions, actions or concepts specific to each image.


In addition, the answer also connects 
the most significant\footnote{Formally, an aspect is as significant as the specificity of the information it contains.} aspects of 
the images. 
Other possible answers like ``nature'' or ``outdoors'' do not demonstrate such properties. They are too general. In essence, image riddles is a challenging task that not only tests our ability to detect visual items in a set of images, but also tests our knowledge and our ability to think and reason. 


   
  Based on the above analysis, 
   we argue that a system should have the following capabilities to answer Image Riddles appropriately: i) the ability to \textit{detect} and locate the objects, regions, and their properties; ii) the ability to recognize \textit{actions}; iii) the ability to \textit{infer} concepts 
from the detected words; and iv) the ability to rank a concept (described in words) based on its relative appropriateness; in other words, the ability to  \textit{reason} with and \textit{process} background or commonsense knowledge about the semantic similarity and relations between words and phrases.  
These capabilities, in fact, are also desired of any automated system that aims to understand a scene and answer questions about it. For example, in VQA dataset \cite{VQA}, ``Does this man have children'', ``Is this a vegetarian Pizza?'' are some such examples, where one needs explicit commonsense knowledge. 
 
   

 These riddles can be thought of as a visual counterpart to IQ test question types such as sequence filling $( x_1, x_2, x_3, ?)$ and analogy solving $(x_1 : y_1 :: x_2 : \  ?)$\footnote{Examples are: word analogy tasks (male : female :: king : ?); numeric sequence filling tasks: $(1,2,3,5,?)$.}  where one needs to find commonalities between items. This task is different from traditional VQA, as in VQA the queries provide some clues regarding what to look for in the image in question. Most riddles in this task require both superior detection and reasoning capabilities, whereas a large percentage (of questions) of the traditional VQA dataset tests system's detection capabilities. This task differs from both VQA and Captioning in that this task requires analysis of multiple images. While video analysis may require analysis of multiple images, this task of  ``image riddles'' focuses on analysis of seemingly different images.
 
Hence, this task of Image Riddles is simple to explain; shares similarities with well-known and pre-defined types of IQ questions and it requires a combination of vision and reasoning capabilities. In this paper, we introduce a promising approach in tackling the problem. 

In our approach, we first use state-of-the-art Image Classification techniques \cite{clarifai} to get the top identified class-labels 
from each image. Given these probabilistic detections, we use the knowledge of connections and relations of these words to infer a set of most probable words (or phrases).  We use ConceptNet 5 \cite{Liu:2004:CMP:1031314.1031373} as the source of commonsense and background knowledge that encodes the relations between words and short phrases using a structured graph.  Note, the possible range of candidates are the entire {\bf vocabulary} of ConceptNet 5 (roughly 0.2 million). For representation and reasoning with this huge probabilistic knowledge we use the Probabilistic Soft Logic (PSL) \cite{kimmig2012short,bach2013hinge} framework\footnote{PSL is shown to be a powerful framework for high-level Computer Vision tasks like Activity Detection \cite{london2013collective}.}. Given the inferred words for each image, we then infer the final set of answers for each riddle. 

   Our \textbf{contributions} are threefold: i) we introduce the 3K Image Riddles Dataset; ii) we present a probabilistic reasoning approach to solve the riddles with reasonable accuracy; iii) our reasoning module inputs detected words (a closed set of class-labels) and \textit{logically} infers all relevant concepts (belonging to a much larger vocabulary).


\vspace{-0.08in}
\section{Related Work} 
The problem of Image Riddles has some similarities to the genre of topic modeling \cite{Blei:2012:PTM:2133806.2133826} and Zero-shot Learning \cite{zerodatalearning}. However, this dataset imposes a few unique challenges: i) the possible set of target labels is the entire Natural Language vocabulary; ii) each image, when grouped with different set of images can map to a different label; iii) almost all the target labels in the dataset are unique (3k examples with 3k class-labels). These challenges make it hard to directly adopt topic model-based or Zero-shot learning-based approaches.

Our work is also related to 
the field of \textbf{Visual Question Answering}.  Very recently, researchers spent a significant amount of efforts on both creating datasets and proposing new models \cite{VQA,malinowski2015ask,gao2015you,ma2015learning}. Interestingly both \cite{VQA} and \cite{gao2015you} adapted MS-COCO \cite{lin2014microsoft} images and created an open domain dataset with human generated questions and answers. Both \cite{malinowski2015ask} and \cite{gao2015you} use recurrent networks to encode the sentence and output the answer. 

Even though some questions from \cite{VQA} and \cite{gao2015you} are very challenging which actually require logical reasoning in order to answer correctly, popular approaches are still hoping to learn the direct signal-to-signal mapping from image and question to its answer, given a large enough annotated data. The necessity of common-sense reasoning could be easily neglected. Here we introduce the new Image Riddle problem
which is 1) a well-defined cognitively challenging task that requires both vision and reasoning capability, 2) it is impossible to model the problem as direct signal-to-signal mapping, due to the data sparsity 
and 3) system's performance could still be bench-marked automatically for comparison. All these qualities make our Image Riddle dataset a good testbed for vision and reasoning research.


\section{Background}
  In this Section, we briefly introduce the different techniques and Knowledge Sources used in our system.
  
\subsection{Probabilistic Soft Logic (PSL)}
PSL is a recently proposed framework for Probabilistic Logic 
\cite{kimmig2012short,bach2013hinge}. A PSL model is defined using a set of weighted if-then rules in first-order logic. 
  
  Let $\bm{C}=(C_1,...,C_m)$ be such a collection where each $C_j$ is a disjunction of literals, where each literal is a variable $y_i$ or its negation $\neg y_i$, where $y_i\in \bm{y}$. Let $I_j^{+}$ (resp. $I_j^{-}$ ) be
  the set of indices of the variables that are not negated (resp. negated) in $C_j$. Each $C_j$ can be written as: 
\begin{equation}  
\setlength{\abovedisplayskip}{1pt}
\setlength{\belowdisplayskip}{1pt}
w_j : \land_{i \in I_j^{-}} y_i \rightarrow \lor_{i \in I_j^{+}} y_i
\end{equation}
or equivalently, $w_j: \lor_{i \in I_j^{-}} (\neg y_i) \bigvee \lor_{i \in I_j^{+}} y_i$. 
  Each rule $C_j$ is associated with a non-negative weight $w_j$. PSL relaxes the boolean truth values of each ground atom $a$ (constant term or predicate with all variables replaced by constants) to the the interval [0, 1], denoted by $I(a)$. To compute soft truth
values for logical formulas, Lukasiewicz’s relaxation \cite{klir1995fuzzy} of conjunctions ($\land$), disjunctions ($\lor$) and negations ($\neg$) 
is used :
\begin{equation}
\setlength{\abovedisplayskip}{1pt}
\setlength{\belowdisplayskip}{1pt}
\begin{aligned}
I(l_1\land l_2) = max\{0,I(l_1)+I(l_2)-1\}\\
I(l_1\lor l_2) = min\{1,I(l_1)+I(l_2)\}\\
I(\neg l_1) = 1-I(l_1)
\end{aligned}
\end{equation}

In PSL, the ground atoms are considered as random variables and the distribution is modeled using {\bf Hinge-Loss Markov Random Field}, which is defined as follows:
\begin{definition}
 Let $\bm{y}$ and $\bm{x}$ be two vectors of $n$ and ${n'}$ random variables respectively, over the domain $D =[0,1]^{n+{n'}}$. The feasible set $\tilde{D}$ is a subset of $D$, defined as:
{\small
\begin{equation*}
\tilde{D} = \big\{ (\bm{y},\bm{x}) \in D \big\rvert \substack{c_k(\bm{y},\bm{x}) = 0, \forall k \in \mathcal{E}\\ c_k(\bm{y},\bm{x}) \leq 0, \forall k \in \mathcal{I}} \big\}
\end{equation*} 
}
where $c=(c_1,...,c_r)$ are linear constraint functions associated with the index sets $\mathcal{E}$ and $\mathcal{I}$ denoting equality and inequality constraints.
A \textit{Hinge-Loss Markov Random Field} $\mathbb{P}$ 
is a probability density, defined as: if $(\bm{y},\bm{x}) \notin \tilde{D}$, then $\mathbb{P}(\bm{y}|\bm{x})=0$; if $(\bm{y},\bm{x}) \in \tilde{D}$, then:
{\small
\begin{equation}
\begin{aligned}
\mathbb{P}(\bm{y}|\bm{x}) =  \frac{1}{Z(\bm{w},\bm{x})}exp(-f_{\bm{w}}(\bm{y},\bm{x}))\\
\end{aligned}
\end{equation}
}
where
$
Z(\bm{w},\bm{x}) = \int_{\bm{y}| (\bm{y},\bm{x}) \in \tilde{D}} exp(-f_{\bm{w}}(\bm{y},\bm{x})) d\bm{y}.
$

The hinge-loss energy function $f_{\bm{w}}$ is defined as:
$
f_{\bm{w}}(\bm{y},\bm{x}) = \sum\limits_{j=1}^{m} w_j(max\{\bm{l_j}(\bm{y},\bm{x}),0\})^{p_j},
$
where  
$w_{j}$'s are non-negative free parameters and 
$\bm{l_j}(\bm{y},\bm{x})$ are linear constraints over $\bm{y},\bm{x}$ and $p_j=\{1,2\}$.

The final inference objective of HL-MRF is:
{\small
\begin{equation}
\setlength{\abovedisplayskip}{0pt}
\mathbb{P}(\bm{y}|\bm{x}) \equiv \argmin_{\bm{y}\in [0,1]^n} \sum_{j=1}^m w_j (max\{\bm{l_j}(\bm{y},\bm{x}),0\})^{p_j}
\end{equation}
}
\end{definition}

In PSL, each logical rule $C_j$ in the database $\bm{C}$ is used to define $\bm{l_j}(\bm{y},\bm{x})$ i.e. the linear constraints over $(\bm{y},\bm{x})$.  Given
a set of weighted logical formulas, PSL builds a graphical model defining a probability distribution over the continuous space of values of the random
variables in the model.

The final optimization problem is defined in terms of ``distance to satisfaction''.
For each rule $C_j \in \bm{C}$ 
this distance to satisfaction  is measured using the term $w_j\times max\big\{ 1- \sum_{i \in I_j^{+}} y_i - \sum_{i \in I_j^{-}} (1-y_i),0\big\}$. This encodes the penalty to the system if a rule is not satisfied. 
 The final optimization problem becomes:
 {\small
\begin{equation}
\argmin_{\bm{y}\in [0,1]^n} \sum_{C_j \in \bm{C}} w_j\text{ }max\big\{ 1- \sum_{i \in I_j^{+}} y_i - \sum_{i \in I_j^{-}} (1-y_i),0\big\}
\end{equation}
}
\vspace*{-0.1in}
\subsection{ConceptNet}
 ConceptNet \cite{speer2012representing}, is a multilingual Knowledge Graph, that encodes commonsense knowledge about the world and is built primarily to assist systems that attempts to understand natural language text. The knowledge in ConceptNet is semi-curated. 
 The nodes (called concepts) in the graph are words or short phrases  written in natural language. The nodes are connected by edges (called assertions) which are labeled with meaningful relations (selected from a well-defined closed set of relation-labels). For example: \textit{(reptile, IsA, animal), (reptile, HasProperty, cold blood)} are some edges. Each edge has an associated confidence score. Also, compared to other knowledge-bases like WordNet, YAGO, NELL \cite{Suchanek:2007:YCS:1242572.1242667,NELL-aaai15}; ConceptNet has a more extensive coverage of English language words and phrases. These properties make this Knowledge Graph a perfect source for the required probabilistic commonsense knowledge.

\subsection{Word2vec}
Word2vec uses the theory of distributional semantics\footnote{The central idea is: ``a word is known by the company it keeps''.} to capture word meanings and produce word embeddings (vectors). The pre-trained word-embeddings  have been successfully used in numerous Natural Language Processing applications and the induced vector-space is known to capture the graded similarities between words with reasonable accuracy \cite{mikolov2013efficient}. Throughout the paper, for word2vec-based similarities, we use the 3 Million word-vectors trained on Google-News corpus \cite{mikolov2013efficient}.

\vspace{-0.08in}
\section{Approach}
Given a set of images (in our case four:  $\{\mathcal{I}_1,\mathcal{I}_2,\mathcal{I}_3,\mathcal{I}_4\}$), the objective is to determine a set of ranked words ($T$) based on how well the word semantically connects these image.
In this work, we present an approach that uses Probabilistic Reasoning on top of a probabilistic Knowledge Base (ConceptNet). It also uses additional  semantic knowledge of words from Word2vec. Using these knowledge sources, we predict the answers to the riddles. 
 
\subsection{Outline of our Framework}
\vspace{-0.1in}
  \begin{algorithm}
  {\fontsize{8}{8}\selectfont
  \caption{Solving Riddles}\label{algo:searchConcepts}
  \begin{algorithmic}[1]
  \Procedure{UnRiddler}{$\mathcal{I}=\{\mathcal{I}_1,\mathcal{I}_2,\mathcal{I}_3,\mathcal{I}_4\},\mathcal{K}_{cnet}$}
  \For{ $\mathcal{I}_k\in \mathcal{I}$}
 	 \State $\tilde{P}(\bm{S_k}|\mathcal{I}_k)$ = getClassLabelsNeuralNetwork($\mathcal{I}_k$).
 	 \For{$s\in \bm{S_k}$}
 	 \State $\bm{T_{s}},W_m(s,\bm{T_{s}})$ = retrieveTargets($s,\mathcal{K}_{cnet}$); \Comment $W_m(s,t_j) = sim(s,t_j) \forall t_j \in \bm{T_{s}}$
 	 \EndFor
 	  \State $\bm{T_k}$ = rankTopTargets($\tilde{P}(\bm{S_k}|\mathcal{I}_k),\bm{T_{S_k}},W_m$);
 	 \State $I(\hat{T}_k)$ = inferConfidenceStageI($\bm{T_k},\tilde{P}(\bm{S_k}|\mathcal{I}_k)$).
  \EndFor
  \State $I(T)$ = inferConfidenceStageII($[\bm{\hat{T}_k}]_{k=1}^4,[\tilde{P}(\bm{S_k}|\mathcal{I}_k)]_{k=1}^4$).
  \EndProcedure
  \end{algorithmic}
  }
  \end{algorithm} 
\vspace{-0.1in}

\begin{figure}[!htpb]
      \centering
      \includegraphics[height=0.11\textheight,width=\textwidth]{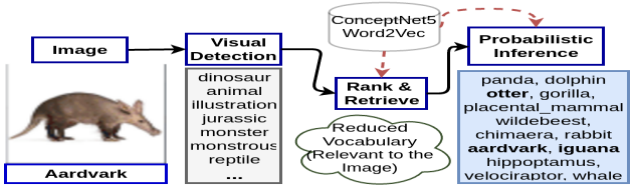}
      \caption{{\small An overview of the framework followed for each Image; demonstrated using an example image of an \textit{aardvark} (resembles animals such as tapir, ant-eater). We run  a similar pipeline for each image and then infer final results using a final Probabilistic Inference Stage (Stage II).}}
      \label{fig:aardvark}
 \end{figure}

 As outlined in algorithm 1, for each image $\mathcal{I}_k$ (here, $k \in \{1,...,4\}$), we follow three stages to infer related words and phrases: i) Image Classification: we get top class labels and the confidence from Image Classifier ($\bm{S_k}, \tilde{P}(\bm{S_k}|\mathcal{I}_k)$), ii) Rank and Retrieve: using these labels and confidence scores, we rank and retrieve top related words from ConceptNet ($\mathcal{K}_{cnet}$), iii) Probabilistic Reasoning and Inference (Stage I): using the labels ($\bm{S_k}$) and the top related words ($\bm{T_k}$), we design an inference model to logically infer final set of words ($\bm{\hat{T}_k}$) for each image.  Lastly, we use another probabilistic reasoning model (Stage II) on the combined set of inferred words (\textit{targets}) from all images in a riddle. This model assigns the final confidence scores on the combined set of targets ($T$). The pipeline followed for each image is depicted with an example in Figure \ref{fig:aardvark}.
 \vspace*{-0.05in}

  \subsection{Image Classification}
Neural Networks trained on ample source of images and numerous image classes has been very effective. Studies have found that convolutional neural networks (CNN) can produce near human level image classification accuracy \cite{krizhevsky2012imagenet}, and related work has been used in various visual recognition 
tasks such as scene labeling \cite{farabet2013learning} and object recognition \cite{girshick2014rich}. To exploit these advances, we use the state-of-the-art class detections provided by the Clarifai API \cite{clarifai} and the Deep Residual Network Architecture by \cite{he2015deep} (using the trained ResNet-200 model). For each image ($\mathcal{I}_k$) we use top 20 detections ($\bm{S_k}$). Let us call these detections as $seeds$. An example is provided in the Figure \ref{fig:aardvark}. Each detection is accompanied with the classifier's confidence score ($\tilde{P}(\bm{S_k}|\mathcal{I}_k)$).

 \subsection{Rank and Retrieve Related Words}

 
Our goal is to logically infer words or phrases that represent (higher or lower-level) concepts that can best explain the co-existence of the \textit{seeds} in a scene.  
 Say, for ``hand'' and ``care'', implied words could be ``massage'', ``ill'',  ``ache'' etc. For ``transportation'' and ``sit'', implied words/phrases could be ``sit in bus'', ``sit in plane'' etc. The reader might be inclined to infer other concepts. However, to ``infer'' is to derive ``logical'' conclusions. Hence, we prefer the concepts which shares strong explainable connections with the seed-words. 
 
 A logical choice would be traversing a knowledge-graph like ConceptNet 
 and find the common reachable nodes from these \textit{seeds}.
  As this is computationally quite infeasible, we use the association-space matrix representation of ConceptNet, where the words are represented as vectors. The similarity between two words approximately embodies the strength of the connection over all paths connecting the two words in the graph. 
  We get the top similar words for each \textit{seed}, approximating the reachable nodes.

\subsubsection{Retrieve Related Words For a Seed}

 \textbf{Visual Similarity}: We observe that, for objects, the ConceptNet-similarity gives a poor result (See Table \ref{tab:men}). So, we define a metric called {\bf visual similarity}. Let us call the similar words as \textit{targets}. In this metric, we represent the seed and the target as vectors. To define the dimensions, for each \textit{seed}, we use a set of relations (HasA, HasProperty, PartOf and MemberOf). We query ConceptNet to get the related words (say, \textit{W1,W2,W3...}) under such relations for the seed-word and its superclasses. 
Each of these  relation-word pairs  (i.e. \textit{HasA-W1,HasA-W2,PartOf-W3,...}) becomes a separate dimension. The values for the seed-vector are the weights assigned to the assertions. For each \textit{target}, we query ConceptNet  and populate the target-vector using the edge-weights for the dimensions defined by the seed-vector.

 To get the top words using visual similarity, we use the cosine similarity of the seed-vector and the target-vector to re-rank the top $10000$ retrieved similar target-words using ConceptNet-similarity. For abstract seed-words, we do not get any such relations and we use the ConceptNet similarity directly.
 \vspace*{-0.1in}
\begin{table}[!htb]
\centering
\caption{Top 10 similar Words for ``Men''. More in appendix.}
\label{tab:men}
\resizebox{\columnwidth}{!}{%
\begin{tabular}{|c|c|c|}
\hline
\textbf{ConceptNet} & \textbf{Visual Similarity} & \textbf{word2vec} \\ \hline\hline
\begin{tabular}[c]{@{}c@{}}man, merby, misandrous,\\ philandry, male\_human,\\ dirty\_pig, mantyhose,\\ date\_woman,guyliner,manslut\end{tabular} & \begin{tabular}[c]{@{}c@{}}priest, uncle, guy,\\ geezer, bloke, pope,\\ bouncer, ecologist,\\ cupid, fella\end{tabular} & \begin{tabular}[c]{@{}c@{}}women, men, males,\\ mens, boys, man, female,\\ teenagers,girls,ladies\end{tabular}\\
\hline
\end{tabular}
}
\vspace*{-0.07in}
\end{table}
\vspace*{-0.05in}
  Table \ref{tab:men} shows the top similar words using ConceptNet, word2vec and visual-similarity for the word ``men".
Moreover, the ranked list based on visual-similarity ranks \textit{boy, chap,  husband, godfather, male\_person, male} in the ranks $16$ to $22$.

  \textbf{Formulation:} For each seed ($s$), we get the top words ($\bm{T_s}$)  from ConceptNet using the visual similarity metric  and the similarity vector $W_m(s,\bm{T_s})$. Together for an image, these constitute $\bm{T_{S_k}}$ and the matrix $\matr{W_m}$, where $\matr{W_m(s_i,t_j)} = sim_{vis}(s_i,t_j) \forall s_i \in S_k, t_j \in \bm{T_{S_k}}$. Next we describe the defined similarity metric.
  
A large percentage of the error in Image Classifiers  are due to visually similar (or semantically similar) objects or objects from the same category \cite{hoiem2012diagnosing}. In such cases,  we use this visual similarity metric to predict the possible visually similar objects and then use an inference model to infer the actual object.


\vspace{-0.08in}
\subsubsection{Rank Targets}

  We use $\tilde{P}(\bm{S_k}|\mathcal{I}_k)$ as an approximate vector representation for the image, in which the seed-words are the dimensions. The columns of $\matr{W_m}$ provides vector representations for the target words ($t \in \bm{T_{S_k}}$) 
  in the space. We calculate cosine similarities for each target with such a image-vector and then re-rank the targets. 
  We consider the top $\bm{\theta}_{\#t}$ targets and we call it $\bm{T_k}$.

\subsection{Probabilistic Reasoning and Inference}

\subsubsection{PSL Inference Stage I} 
  Given a set of candidate \textit{targets} $\bm{T_k}$ and a set of weighted \textit{seeds} ($\bm{S_k},\tilde{P}(\bm{S_k}|\mathcal{I}_k)$), we build an inference model to infer a set of most probable \textit{targets} ($\bm{\hat{T}_k}$). We model the joint distribution using PSL as this formalism adopts Markov Random Field which obeys the properties of Gibbs Distribution. In addition, a PSL model is declared using rules. Given the final answer from the system, the set of satisfied (grounded) rules show the logical connections between the detected words and the final answer, which demonstrates the system's explainability. 
  
\begin{figure}[!htpb]
      \centering
      \includegraphics[height=0.1\textheight]{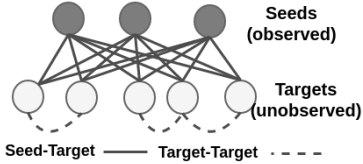}
      \caption{{\small Joint Modeling of seeds and targets, depicted as a Undirected Graphical Model. We define the seed-target and target-target potentials using PSL rules. We connect each seed to each target and the potential depends on their similarity and the target's popularity bias. We connect each target to $\bm{\theta}_{t\text{-}t}$ (1 or 2) maximally similar targets. The potential depends on their similarity.}}
      \label{fig:psl}
 \end{figure}

  The PSL model can be best depicted as an Undirected Graphical Model involving \textit{seeds} and \textit{targets}, as given in Figure \ref{fig:psl}. 
  \vspace*{-0.07in}

  \textbf{Formulation:} Using PSL, we add two sets of rules: i) to define seed-target potentials, we add rules of the form $wt_{ij}: s_{ik} \rightarrow t_{jk}$ for each word $s_{ik} \in \bm{S_k}$ and target $t_{jk} \in \bm{T_k}$; ii) to define target-target potentials, for each target $t_{jk}$, we take the most similar $\bm{\theta}_{t\text{-}t}$ targets ($T^{max}_{j}$). For each target $t_{jk}$ and each $t_{mk} \in T^{max}_{j}$, we add two rules $wt_{jm}: t_{jk} \rightarrow t_{mk}$ and $wt_{jm}: t_{mk} \rightarrow t_{jk}$. Next, we describe the choices in detail.
  
  i) From the perspective of optimization, the rule $wt_{ij}: s_{ik} \rightarrow t_{jk}$ adds the term $wt_{ij}*max\{I(s_{ik})-I(t_{jk}),0\}$ to the objective. This means that if confidence score of the target $t_{jk}$ is not greater than $I(s_{ik})$ (i.e. $\tilde{P}(\bm{S_k}|\mathcal{I}_k)$), then the rule is not satisfied and we penalize the model by $wt_{ij}$ times the difference between the confidence scores. We add the above rule for seeds and targets for which the combined weighted similarity exceeds certain threshold $\bm{\theta}_{sim,psl1}$. 
  
We encode the commonsense knowledge of words and phrases obtained from different knowledge sources into the weights of these rules $wt_{ij}$. Both the knowledge sources are considered because ConceptNet embodies commonsense knowledge and word2vec encodes word-meanings. 
It is also important that the inference model is not biased towards more popular targets (i.e. abstract words or words too commonly used/detected in corpus).   
 We compute eigenvector centrality score ($\mathbb{C}(.)$) for each word in the context of  ConceptNet (a network of words and phrases).
 Higher $\mathbb{C}(.)$ indicates higher connectivity 
 of a word in the graph. This yields a higher similarity score to many words and might give an unfair bias to this \textit{target} in the inference model. Hence, the higher the $\mathbb{C}(.)$, the word provides less specific information for an image.
Hence, the weight becomes
 {\small
  \begin{equation}\label{eq:10}
  \begin{aligned}
  wt_{ij} &= \bm{\theta}_{\alpha_1} * sim_{cn}(s_{ik}, t_{jk}) + \\
  &\bm{\theta}_{\alpha_2} * sim_{w2v}(s_{ik}, t_{jk}) 
   + 1/\mathbb{C}(t_{jk}),
  \end{aligned}
  \end{equation} 
  }
where $sim_{cn}(.,.)$ is the normalized ConceptNet-based similarity.  $sim_{w2v}(.,.)$ is the normalized word2vec similarity of two words and $\mathbb{C}(.)$ is the eigenvector-centrality score of the argument in the ConceptNet matrix.  
  
   ii) To model dependencies among the targets, we observe that if two concepts $t_1$ and $t_2$ are very similar in meaning, then a system that infer $t_1$ should infer $t_2$ too, given the same set of observed words. Therefore, the two rules $wt_{jm}: t_{jk} \rightarrow t_{mk}$ and $wt_{jm}: t_{mk} \rightarrow t_{jk}$ are designed to force the confidence values of $t_{jk}$ and $t_{mk}$ to be as close to  each other as possible. $wt_{jm}$ is the same as Equation \ref{eq:10} without the penalty for popularity.

The combined PSL model inference objective becomes:
{\small
\begin{equation}
\begin{aligned}
  \argmin_{I(\bm{T_{k}})\in [0,1]^{|T_{k}|}} \sum_{s_{ik} \in \bm{S_k}} \sum_{t_{jk} \in \bm{T_{k}}} wt_{ij}\text{ } max\big\{I(s_{ik})-I(t_{jk}),0\big\} + \\
  \sum_{t_{jk} \in \bm{T_{k}}} \sum_{t_{mk} \in T^{max}_{j}} wt_{jm}\text{ }\Big\{max\big\{I(t_{mk})-I(t_{jk}),0\big\} + \\
  max\big\{I(t_{jk})-I(t_{mk}),0\big\} \Big\}. \nonumber
\end{aligned}
\end{equation}
}
   To let the targets compete against each other, 
   we add a constraint on the sum of the confidence scores of the targets i.e. $\sum_{j:t_{jk} \in \bm{T_{k}}} I(t_{jk}) \leq \bm{\theta}_{sum1} $. Here $\bm{\theta}_{sum1} \in \{1,2\} $ and  $I(t_{jk}) \in [0,1]$.
  As a result of this model, we get an inferred reduced set of targets $[\bm{\hat{T}_{k}}]_{k=1}^4$.

\subsubsection{PSL Inference Stage II} 

To learn the most probable set of common targets jointly, we consider the \textit{targets} and the \textit{seeds} from all images together. Assume that the \textit{seeds} and the \textit{targets} are nodes in a knowledge-graph. Then, the most appropriate target-nodes 
  should observe similar  properties as an appropriate answer to the riddle: i) 
a target-node should be connected to the high-weight seeds in an image i.e. should relate to the important aspects of the image; ii) a target-node should be connected to seeds from all images.

 \textbf{Formulation:}   
 Here, we use the rules $wt_{ij}: s_{ik} \rightarrow t_{jk}$ for each word $s_{ik} \in \bm{S_k}$ and target $t_{jk} \in \bm{\hat{T}_{k}}$ for all $k\in \{1,2..,4\}$.  
  To let the set of targets compete against each other, we add the constraint  $\sum_{k=1}^4 \sum_{j:t_{jk} \in \bm{\hat{T}_{k}}} I(t_{jk}) \leq \bm{\theta}_{sum2} $. Here $\bm{\theta}_{sum2}=1 $ and  $I(t_{jk}) \in [0,1]$.
  
  To minimize the penalty for each rule, the optimal solution will try to maximize the confidence score of $t_{jk}$. To minimize the overall penalty, it should maximize the confidence scores of those targets which will satisfy most of the rules (or rules with maximum total weight). As the summation of confidence scores is bounded, only a few top inferred targets should have non-zero confidence.

\section{Experiments and Results}
  In this section, we provide the results of the validation experiments of the newly introduced Image Riddle dataset, followed by empirical evaluation of the proposed approach against vision-only baselines.
  
\vspace{-0.05in}
\subsection{Dataset Validation and Analysis}
\vspace*{-0.05in}
 We have collected a set of $3333$ riddles from the internet (puzzle websites). Each riddle has $4$ images ($66\times66$, 6KB in size) and a groundtruth label associated with it. To verify the groundtruth answers, we define the metrics: i) ``correctness'' - how correct and appropriate the answers are, and ii) ``difficulty'' - how difficult are the riddles. We conduct an Amazon Mechanical Turker-based evaluation. We ask them to rate the correctness from 1-6\footnote{1: Completely gibberish, incorrect, 2: relates to one image, 3 and 4: connects two and three images respectively, 5: connects all 4 images, but could be a better answer, 6: connects all images and an appropriate answer.}. The ``difficulty'' is rated from 1-7\footnote{These gradings are adopted from VQA AMT instructions \cite{VQA}. 1: A toddler can solve it (ages:3-4), 2: A younger child can solve it (ages:5-8), 3: A older child can solve it (ages:9-12), 4: A teenager can solve it (ages:13-17), 5: An adult can solve it (ages:18+), 6: Only a Linguist (one who has above-average knowledge about English words and the language in general) can solve it, 7: No-one can solve it.}. 
 According to the Turkers, the mean correctness rating is 4.4 (with Standard Deviation 1.5). The ``difficulty'' ratings show the following distribution: toddler ($0.27\%$), younger child ($8.96\%$), older child ($30.3\%$), teenager ($36.7\%$), adult ($19\%$), linguist ($3.6\%$), no-one ($0.64\%$). In short, the average age to answer the riddles seems to be closer to \textbf{13-17yrs}. Also, few of these ($4.2\%$) riddles seem to be incredibly hard. Interestingly, the average age perceived reported for the recently proposed VQA dataset \cite{VQA} is \textbf{8.92 yrs}. Although, this experiment measures ``the turkers' perception of the required age'', one can conclude that the riddles are comparably harder.

\subsection{System Evaluation}
  The presented approach suggests the following hypotheses that requires empirical tests: I) the proposed approach (and their variants) attain reasonable accuracy in solving the riddles; II) the individual stages of the framework improves the final inference accuracy of the answers. In addition, we also experiment to observe the effect of using commercial classification methods like Clarifai against a published state-of-the-art Image Classification method. 
  

\vspace{-0.05in}
\subsubsection{Systems}
 We propose several variations of the proposed approach and compare them with a simple vision-only baseline (hypothesis I). 
 We introduce an additional Bias-Correction stage after the Image Classification, which aims to re-weight the detected seeds using additional information from other images.
 The variations then, are created to test the effects of varying the Bias-Correction  stage and the effects of the individual stages of the framework on the final accuracy (hypothesis II). We also vary the initial Image Classification Method (Clarifai, Deep Residual Network).
 
 \textbf{Bias-Correction:} We experimented with two variations: i) greedy bias-correction and ii) no bias-correction.    
   We follow the intuition that the re-weighting of the seeds of one image can be influenced by the others\footnote{A person would often skim through all the images at one go and will try to come up with the aspects that needs more attention.}. To this end, we develop the ``GreedyUnRiddler'' (\textbf{GUR}) approach.  In this approach, we consider all of the images together to dictate the new weight of each seed. Take image $\mathcal{I}_k$ for example. To re-weight seeds in $\bm{S_k}$, we calculate the weights using the following equation: $\tilde{W}(s_k) = \frac{\sum_{j \in {1,..4}} sim_{cosine}(V_{s_k,j}, V_j)}{4.0}$.
    $V_j$ is vector of the weights assigned $\tilde{P}(\bm{S_j}|\mathcal{I}_j)$ i.e. confidence scores of each seed in the image. Each element of $V_{s_k,j}[i]$ is the ConceptNet-similarity score between the seed $s_k$ and  $s_{i,j}$ i.e. the $i^{th}$ seed of the $j^{th}$ image. The re-weighted seeds ($\bm{S_k},\tilde{W}(\bm{S_k})$) of an image are then passed through the rest of the pipeline to infer the final answers. 
   
   In the original pipeline (``UnRiddler'',in short \textbf{UR}), we just normalize the weights of the seeds and pass on to the next stage. We experiment with another variation (called BiasedUnRiddler or \textbf{BUR}), the results of which are included in appendix, as \textbf{GUR} achieves the best results.
  
  \textbf{Effect of Stages:} We observe the accuracy after each stage in the pipeline (\textbf{VB}: Upto Bias Correction, \textbf{RR}: Upto Rank and Retrieve stage, \textbf{All}: The entire Pipeline). For \textbf{VB}, we use the normalized weighted seeds, get the weighted centroid vector over the word2vec embeddings of the seeds for each image. Then we obtain the mean vector over these centroids. The top similar words from the word2vec vocabulary to this mean vector, constitutes the final answers. For \textbf{RR}, we get the mean vector over the top predicted targets for all images. Again, the most similar words from the word2vec vocabulary constitutes the answers.
  
  \textbf{Baseline:} 
  We create Vision-only Baselines. 
   We directly use the class-labels and the confidence scores predicted using a Neural Network-based Classifier. 
   For each image, we calculate the weighted centroid of the  word2vec embeddings of these labels 
    and the mean of these centroids for the 4 images. For the automatic evaluation we use this centroid and for the human evaluation, we use the most similar word to this vector, from the word2vec vocabulary. The Baseline performances are listed in Table \ref{tab:accuracy} in the \textbf{VB+UR} cells.


\vspace*{-0.05in}
\subsubsection{Experiment I: Automatic Evaluation}
\vspace*{-0.05in}
   We evaluate the performance of the proposed approach on the 3333 Image Riddles dataset using both automatic and Amazon Mechanical Turker (AMT)-based evaluations.  
   
   As an evaluation metric, we use word2vec similarity measure. 
   An answer to a riddle may have several semantically similar answers. Hence it is reasonable to use such a metric. For each riddle, we calculate the maximum similarity between the groundtruth and top 10 detections from an approach. To calculate phrase similarities, we use \texttt{n\_similarity} method of the \texttt{gensim.models.word2vec} package. The average of such maximum similarities is reported in percentage form.


 \begin{table}[!htp]
\centering
\vspace*{-0.09in}
\caption{\small Accuracy on the Image Riddle Dataset. Pipeline variants (VB, RR and All) are combined with Bias-Correction stage variants (GUR, UR). All values are in percentage form. (\textbf{*}- Best, $\dagger$ - Baselines).}
\label{tab:accuracy}
\resizebox{\columnwidth}{!}{%
\fontsize{8}{9}\selectfont
\begin{tabular}{cccccc}
\cline{3-6}
\multicolumn{1}{l}{}      &                         & \multicolumn{2}{c}{GUR}                             & \multicolumn{2}{c}{UR}                                       \\ \cline{3-6} 
\multicolumn{1}{l}{}      &                         & 3.3k                     & 2.8k                     & 3.3k                      & 2.8k                             \\ \hline
\multirow{3}{*}{Clarifai} & VB                      & 65.3                     & 65.36                    & 65                        & 65.3$^\dagger$                   \\ \cline{2-6} 
                          & RR                      & 65.9                     & 65.73                     & 65.9                      & 65.7                             \\ \cline{2-6} 
                          & All                     & \textbf{68.8*}           & \textbf{68.7}            & 68.5                      & \textbf{68.57}                   \\ \hline
\multirow{3}{*}{ResNet}   & \multicolumn{1}{l}{VB}  & \multicolumn{1}{l}{66.8} & \multicolumn{1}{l}{66.4} & \multicolumn{1}{l}{68.3}  & \multicolumn{1}{l}{68$^\dagger$} \\ \cline{2-6} 
                          & \multicolumn{1}{l}{RR}  & \multicolumn{1}{l}{66.3} & \multicolumn{1}{l}{66.2} & \multicolumn{1}{l}{67}  & \multicolumn{1}{l}{66.7}         \\ \cline{2-6} 
                          & \multicolumn{1}{l}{All} & \multicolumn{1}{l}{68.2} & \multicolumn{1}{l}{68.2} & \multicolumn{1}{l}{\textbf{68.53}} & \multicolumn{1}{l}{68.2}         \\ \hline
\end{tabular}
}
\end{table} 
\vspace*{-0.07in}
\vspace*{-0.1in}
\begin{table}[!htb]
\centering
\vspace*{-0.08in}
\caption{A List of parameters $\bm{\theta}$ used in the approach}
\label{my-label}
\resizebox{\columnwidth}{!}{%
\fontsize{7}{8}\selectfont
\begin{tabular}{|l|l|l|}
\hline
$\bm{\theta}_{\#t}$        & Number of Targets                   & 2500 \\ \hline
$\bm{\theta}_{\alpha_1}$   & ConceptNet-similarity Weight        & 1    \\ \hline
$\bm{\theta}_{\alpha_2}$   & word2vec-similarity weight          & 4    \\ \hline
$\bm{\theta}_{t\text{-}t}$ & Number of maximum similar Targets   & 1    \\ \hline
$\bm{\theta}_{sim,psl1}$   & Seed-target similarity Threshold     & 0.8  \\ \hline
$\bm{\theta}_{sum1}$       & Sum of confidence scores in Stage I & 2    \\ \hline
\end{tabular}
}
\end{table}
\vspace*{-0.07in}
To select the parameters in the parameter vector $\bm{\theta}$, We employed a random search on the parameter-space over first 500 riddles over $500$ combinations.  The final set of parameters used and their values are tabulated in Table \ref{my-label}.

   Each of the stage-variants (VB, RR and All) are combined with different variations of the Bias-Correction stage (GUR and UR respectively). The accuracies on all are listed in Table \ref{tab:accuracy}. We provide our experimental results on this $3333$ riddles and $2833$ riddles (barring $500$ riddles we used for the parameter search).

\vspace*{-0.07in}
\subsubsection{Experiment II: Human Evaluation}
\begin{figure}[!htpb]
\vspace*{-0.18in}
     \centering
 \subfloat{\includegraphics[height=0.21\textheight,width=\textwidth]{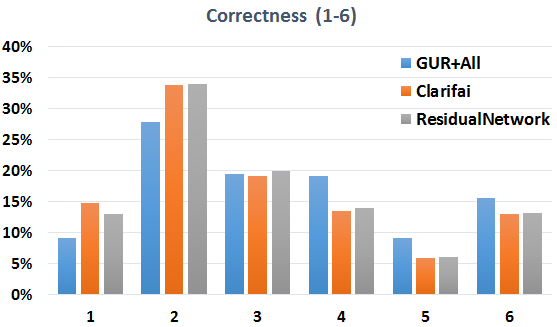}}
      \hfill \subfloat{\includegraphics[height=0.16\textheight,width=\textwidth]{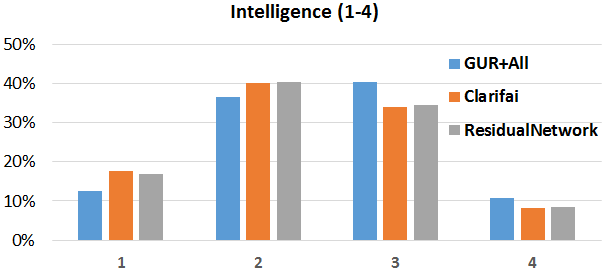} }
       \hfill
      \vspace*{-0.05in}
     \caption{\small AMT Results of The GUR+All (our), Clarifai (baseline 1) and ResidualNet (baseline 2) approaches. Correctness Means are: $2.6\pm1.4$, $2.4\pm1.45$, $2.3\pm1.4$. For Intelligence: $2.2\pm0.87,2\pm0.87,1.8\pm0.8$}.
     \label{fig:amt}
     \vspace*{-0.25in}
\end{figure}

We conduct an AMT-based comparative evaluation of the results of the proposed approach (GUR+All using Clarifai) and two vision-only baselines. We define two metrics: i) ``correctness'' and ii) ``intelligence''. Turkers are presented with a scenario: \textit{We have three separate robots that attempted to answer this riddle. You have to rate the answer based on the correctness and the degree of intelligence (explainability) shown through the answer.}. The correctness is defined as before.  In addition, turkers are asked to rate intelligence in a scale of 1-4\footnote{1: Not intelligent, 2: Moderately Intelligent, 3: Intelligent, 4: Very Intelligent.}. We plot the the percentage of total riddles per each value of correctness and intelligence in Figure \ref*{fig:amt}. In these histograms plots, we expect a increase in the rightmost buckets for the more ``correct'' and ``intelligent'' systems. 

\vspace{-0.15in}
\subsubsection{Analysis}

\vspace*{-0.1in}
\begin{figure*}[htbp!]
     \centering      \subfloat{\includegraphics[width=0.32\textwidth]{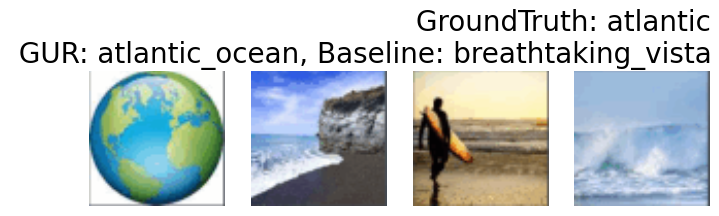}}
      \hfill      \subfloat{\includegraphics[width=0.32\textwidth]{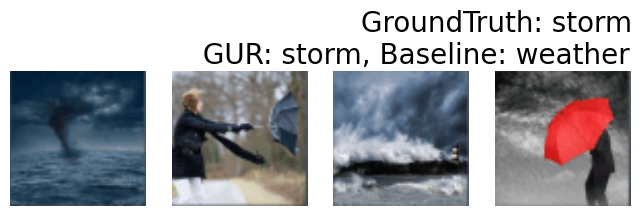}}      
       \hfill      \subfloat{\includegraphics[width=0.32\textwidth]{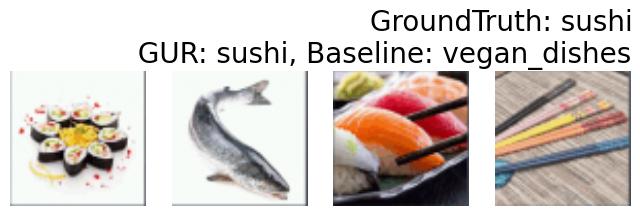}}
      \hfill
\subfloat{\includegraphics[width=0.33\textwidth]{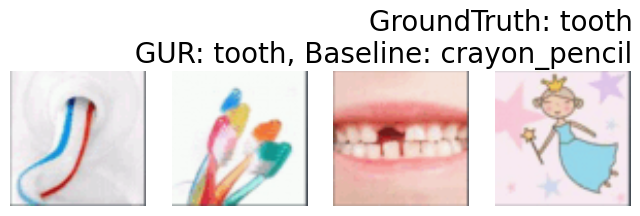} }
      \hfill
\subfloat{\includegraphics[width=0.33\textwidth]{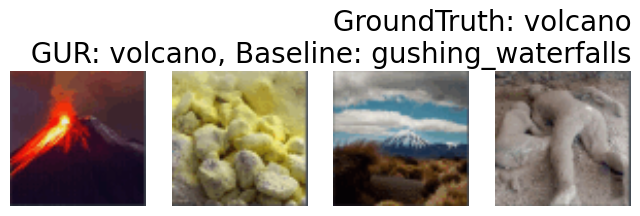}}
\hfill
\subfloat{\includegraphics[width=0.33\textwidth]{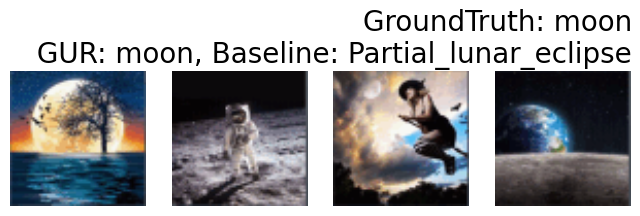}}
\hfill
\subfloat{\includegraphics[width=0.33\textwidth]{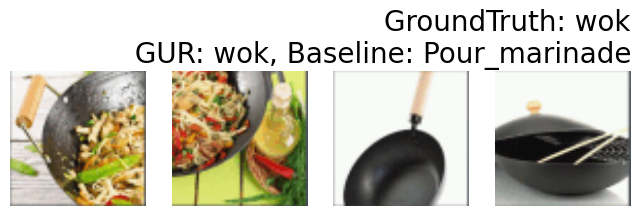}}
\hfill
\subfloat{\includegraphics[width=0.33\textwidth]{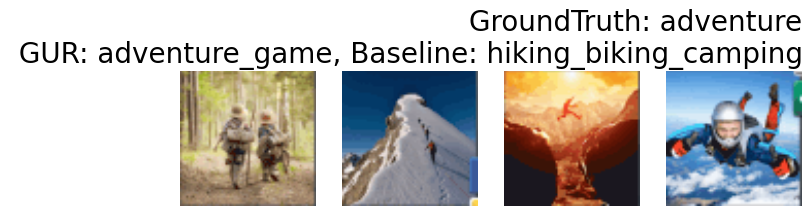} }
\hfill
\subfloat{\includegraphics[width=0.33\textwidth]{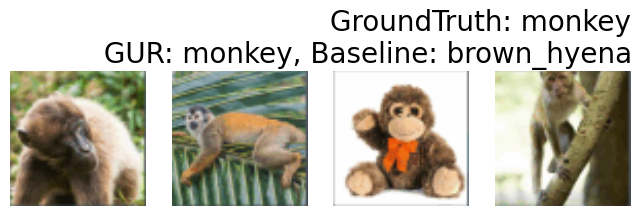}}    
\hfill
\vspace*{-0.02in}
      \subfloat{\includegraphics[width=0.33\textwidth]{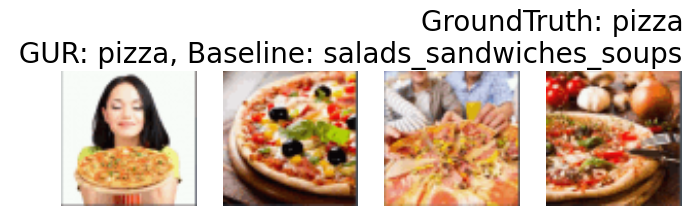} }
\hfill
\subfloat{\includegraphics[width=0.33\textwidth]{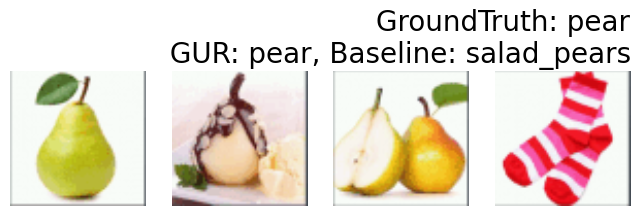}}
\hfill
\subfloat{\includegraphics[width=0.33\textwidth]{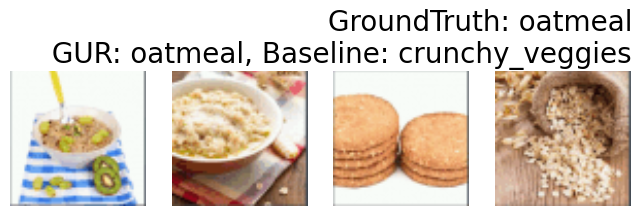}}
\hfill
\vspace*{-0.02in}
\subfloat{\includegraphics[width=0.33\textwidth]{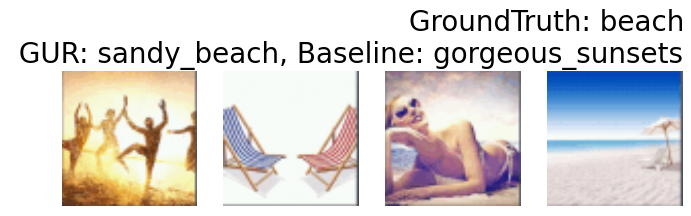}}
\hfill
\subfloat{\includegraphics[width=0.33\textwidth]{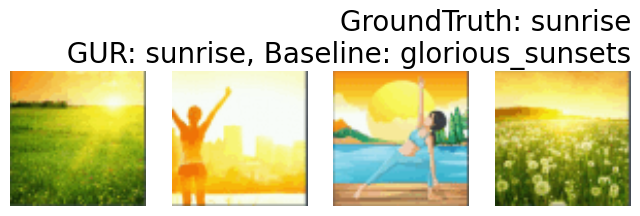}}
\hfill
\subfloat{\includegraphics[width=0.33\textwidth]{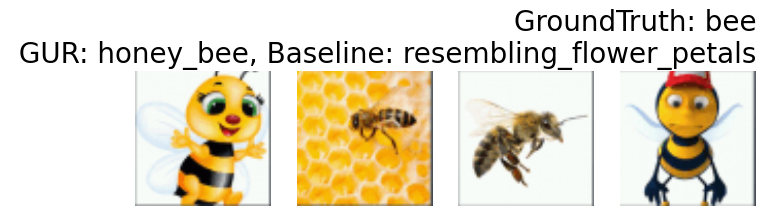}}
\hfill
\vspace*{-0.02in}
\subfloat{\includegraphics[width=0.33\textwidth]{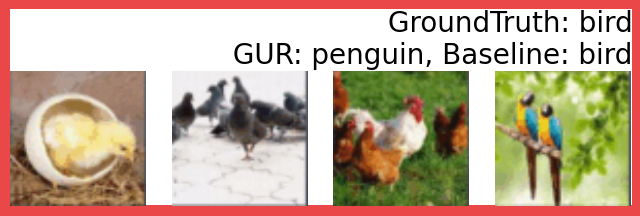}}
\hfill
\subfloat{\includegraphics[width=0.33\textwidth]{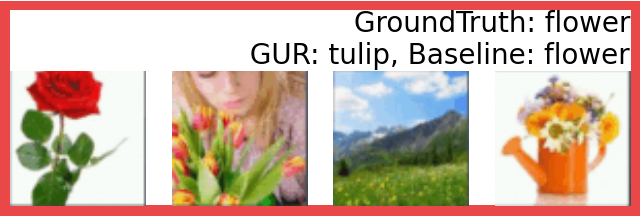}}
\hfill
\subfloat{\includegraphics[width=0.33\textwidth]{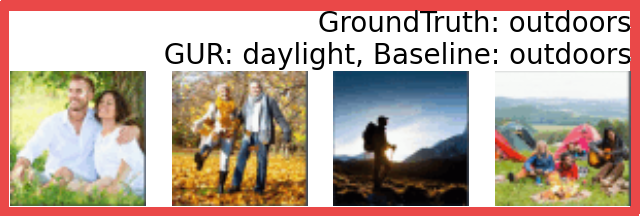}}
      \vspace*{0.06in}
     \caption{Positive and Negative (in red) results of the ``GUR'' approach (\textbf{GUR+All} variant) on some of the riddles. The groudtruth labels, closest label among top 10 from GUR and the Clarifai baseline are provided for all images. For more results, check Appendix and the ImageRiddle website (\href{http://bit.ly/1Rj4tFc}{http://bit.ly/1Rj4tFc}).}
     \label{fig:example_Riddles}
     \vspace*{-0.1in}
\end{figure*} 

  Experiment I shows that the GUR variant (\textbf{GUR+All} in Table \ref{tab:accuracy}) achieves the best results in terms of word2vec-based accuracy. Similar trend is reflected in the AMT-based evaluations (Figure \ref{fig:amt}). Our system has increased the percentage of  puzzles for the rightmost bins i.e. produces more ``correct'' and ``intelligent'' answers for more number of puzzles. The word2vec-based accuracy puts the performance of  ResNet baseline close to that of the GUR variant. However, as evident from Figure \ref{fig:amt}, the AMT evaluation of the correctness shows clearly that the ResNet baseline lags in predicting meaningful answers. Experiment II also includes what the turkers think about the intelligence of the systems that tried to solve the puzzles. This also puts the GUR variant at the top. The above two experiments empirically show that our approach achieves a reasonable accuracy in solving the riddles (Hypothesis I). 
  In table \ref{tab:accuracy}, we observe how the accuracy varies after each stage of the pipeline (hypothesis II). The table shows a jump in the accuracy after the RR stage, which leads us to believe the primary improvement of our approach is attributed to the Probabilistic Reasoning model.  
We also provide our detailed results for the ``GUR'' approach using a few riddles in Figure \ref{fig:example_Riddles}.

\vspace*{-0.05in}
\section{Conclusion and Future Works}
\vspace*{-0.05in}
 In this work, we presented a Probabilistic Reasoning based approach to solve a new class of image puzzles, called ``Image Riddles''. We have collected over 3k such riddles. 
 Crowd-sourced evaluation of the dataset 
 demonstrates the validity of the annotations and the nature of the difficulty of the riddles. 
 We 
 empirically show that our approach improves on vision-only baselines and provides a stronger baseline for future attempts.
 
   The task of ``Image Riddles'' is equivalent to conventional IQ test questions such as analogy solving, sequence filling; which are often used to test human intelligence. This task of ``Image Riddles'' is also in line with the current trend of VQA datasets which require visual recognition and reasoning capabilities. However, it focuses more on the combination of both vision and reasoning capabilities. 
   In addition to the task, the proposed approach introduces a novel inference model to infer related words (from a large vocabulary) given class labels (from a smaller set), using semantic knowledge of words. This method is general in terms of its applications. Systems such as \cite{wu2015value}, which use a collection of high-level concepts to boost VQA performance; can benefit from this approach.
   

{
\bibliographystyle{ieee}
\bibliography{references}

\begin{thebibliography}{10}\itemsep=-1pt

\bibitem{VQA}
S.~Antol, A.~Agrawal, J.~Lu, M.~Mitchell, D.~Batra, C.~L. Zitnick, and
  D.~Parikh.
\newblock Vqa: Visual question answering.
\newblock In {\em International Conference on Computer Vision (ICCV)}, 2015.

\bibitem{bach2013hinge}
S.~Bach, B.~Huang, B.~London, and L.~Getoor.
\newblock Hinge-loss markov random fields: Convex inference for structured
  prediction.
\newblock {\em arXiv preprint arXiv:1309.6813}, 2013.

\bibitem{Blei:2012:PTM:2133806.2133826}
D.~M. Blei.
\newblock Probabilistic topic models.
\newblock {\em Commun. ACM}, 55(4):77--84, Apr. 2012.

\bibitem{brysbaert2014concreteness}
M.~Brysbaert, A.~B. Warriner, and V.~Kuperman.
\newblock Concreteness ratings for 40 thousand generally known english word
  lemmas.
\newblock {\em Behavior research methods}, 46(3):904--911, 2014.

\bibitem{farabet2013learning}
C.~Farabet, C.~Couprie, L.~Najman, and Y.~LeCun.
\newblock Learning hierarchical features for scene labeling.
\newblock {\em Pattern Analysis and Machine Intelligence, IEEE Transactions
  on}, 35(8):1915--1929, 2013.

\bibitem{gao2015you}
H.~Gao, J.~Mao, J.~Zhou, Z.~Huang, L.~Wang, and W.~Xu.
\newblock Are you talking to a machine? dataset and methods for multilingual
  image question answering.
\newblock {\em arXiv preprint arXiv:1505.05612}, 2015.

\bibitem{girshick2014rich}
R.~Girshick, J.~Donahue, T.~Darrell, and J.~Malik.
\newblock Rich feature hierarchies for accurate object detection and semantic
  segmentation.
\newblock In {\em Computer Vision and Pattern Recognition (CVPR), 2014 IEEE
  Conference on}, pages 580--587. IEEE, 2014.

\bibitem{he2015deep}
K.~He, X.~Zhang, S.~Ren, and J.~Sun.
\newblock Deep residual learning for image recognition.
\newblock {\em arXiv preprint arXiv:1512.03385}, 2015.

\bibitem{hoiem2012diagnosing}
D.~Hoiem, Y.~Chodpathumwan, and Q.~Dai.
\newblock Diagnosing error in object detectors.
\newblock In {\em European conference on computer vision}, pages 340--353.
  Springer, 2012.

\bibitem{kimmig2012short}
A.~Kimmig, S.~Bach, M.~Broecheler, B.~Huang, and L.~Getoor.
\newblock A short introduction to probabilistic soft logic.
\newblock In {\em Proceedings of the NIPS Workshop on Probabilistic
  Programming: Foundations and Applications}, pages 1--4, 2012.

\bibitem{klir1995fuzzy}
G.~Klir and B.~Yuan.
\newblock Fuzzy sets and fuzzy logic: theory and applications.
\newblock 1995.

\bibitem{krizhevsky2012imagenet}
A.~Krizhevsky, I.~Sutskever, and G.~E. Hinton.
\newblock Imagenet classification with deep convolutional neural networks.
\newblock In {\em Advances in neural information processing systems}, pages
  1097--1105, 2012.

\bibitem{zerodatalearning}
H.~Larochelle, D.~Erhan, Y.~Bengio, U.~D. Montréal, and M.~Québec.
\newblock Zero-data learning of new tasks.
\newblock In {\em In AAAI}, 2008.

\bibitem{lin2014microsoft}
T.-Y. Lin, M.~Maire, S.~Belongie, J.~Hays, P.~Perona, D.~Ramanan,
  P.~Doll{\'a}r, and C.~L. Zitnick.
\newblock Microsoft coco: Common objects in context.
\newblock In {\em Computer Vision--ECCV 2014}, pages 740--755. Springer, 2014.

\bibitem{Liu:2004:CMP:1031314.1031373}
H.~Liu and P.~Singh.
\newblock Conceptnet - a practical commonsense reasoning tool-kit.
\newblock {\em BT Technology Journal}, 22(4):211--226, Oct. 2004.

\bibitem{london2013collective}
B.~London, S.~Khamis, S.~Bach, B.~Huang, L.~Getoor, and L.~Davis.
\newblock Collective activity detection using hinge-loss markov random fields.
\newblock In {\em Proceedings of the IEEE Conference on Computer Vision and
  Pattern Recognition Workshops}, pages 566--571, 2013.

\bibitem{ma2015learning}
L.~Ma, Z.~Lu, and H.~Li.
\newblock Learning to answer questions from image using convolutional neural
  network.
\newblock {\em arXiv preprint arXiv:1506.00333}, 2015.

\bibitem{malinowski2015ask}
M.~Malinowski, M.~Rohrbach, and M.~Fritz.
\newblock Ask your neurons: A neural-based approach to answering questions
  about images.
\newblock {\em arXiv preprint arXiv:1505.01121}, 2015.

\bibitem{mikolov2013efficient}
T.~Mikolov, K.~Chen, G.~Corrado, and J.~Dean.
\newblock Efficient estimation of word representations in vector space.
\newblock {\em arXiv preprint arXiv:1301.3781}, 2013.

\bibitem{NELL-aaai15}
T.~Mitchell, W.~Cohen, E.~Hruschka, P.~Talukdar, J.~Betteridge, A.~Carlson,
  B.~Dalvi, M.~Gardner, B.~Kisiel, J.~Krishnamurthy, N.~Lao, K.~Mazaitis,
  T.~Mohamed, N.~Nakashole, E.~Platanios, A.~Ritter, M.~Samadi, B.~Settles,
  R.~Wang, D.~Wijaya, A.~Gupta, X.~Chen, A.~Saparov, M.~Greaves, and
  J.~Welling.
\newblock Never-ending learning.
\newblock In {\em Proceedings of the Twenty-Ninth AAAI Conference on Artificial
  Intelligence (AAAI-15)}, 2015.

\bibitem{clarifai}
G.~Sood.
\newblock {\em clarifai: R Client for the Clarifai API}, 2015.
\newblock R package version 0.2.

\bibitem{speer2012representing}
R.~Speer and C.~Havasi.
\newblock Representing general relational knowledge in conceptnet 5.
\newblock 2012.

\bibitem{Suchanek:2007:YCS:1242572.1242667}
F.~M. Suchanek, G.~Kasneci, and G.~Weikum.
\newblock Yago: A core of semantic knowledge.
\newblock In {\em Proceedings of the 16th International Conference on World
  Wide Web}, WWW '07, pages 697--706, New York, NY, USA, 2007. ACM.

\bibitem{wu2015value}
Q.~Wu, C.~Shen, L.~Liu, A.~Dick, and A.~v.~d. Hengel.
\newblock What value do explicit high level concepts have in vision to language
  problems?
\newblock {\em arXiv preprint arXiv:1506.01144}, 2015.

\end{thebibliography}
}
\newpage
\appendix
\section*{Appendices}
\section{BiasedUnRiddler (BUR): A Variation of the BiasCorrection Stage}
 \begin{figure}[!htpb]
      \centering
      \includegraphics[height=0.11\textheight,width=\textwidth]{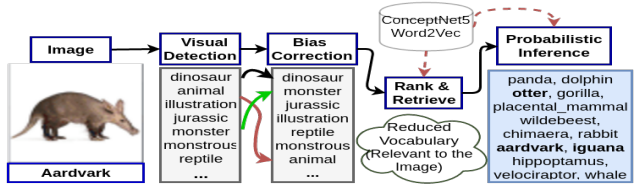}
      \caption{{\small Clarifai detections and results from different stages for the aardvark image (for \textbf{BUR} variant).}}
      \label{fig:aardvark2}
 \end{figure}
In Figure \ref{fig:aardvark2}: \textit{dinosaur, animal} and \textit{reptile} all provide evidence that the image has an animal. Only the word \textit{dinosaur} indicates what kind of animal is in the image. The other words do not add any additional information. Some high-confidence detections also provide erroneous abstract information. Here, the labels \textit{monstrous, monster} are some such detections. Hence, the objective is  to re-weight the seeds so that: i) the more specific seed-words 
should have higher weight than the ones which provide \textit{similar} but more general information; ii) the seeds that are too frequently used or detected in corpus, should be given lower weights.

\textbf{Specificity and Popularity}:   
We compute eigenvector centrality score ($ECS$) for each word in the context of  ConceptNet.  
Higher $ECS$ indicates higher connectivity and  yields a higher similarity score to many words and might give an unfair bias to this \textit{seed} (and words implied by this \textit{seed}) in the inference model. Hence, the higher the $ECS$, the word provides less specific information for an image. Additionally, we use the concreteness rating ($CR$) from \cite{brysbaert2014concreteness}. In this paper, the top $39955$ frequent English words are rated from the scale of 1 (very abstract) to 5 (very concrete). For example, the mean ratings for \textit{monster, animal} and \textit{ dinosaur} are $3.72, 4.61$ and $4.87$ respectively.
 
   \textbf{Problem Formulation}: We formulate the problem as a resource flow problem on a graph. The directed graph $G$ is constructed in the following way: we order the \textit{seeds} based on decreasing centrality scores ($CS$). We compute $CS$ as:
   \begin{equation}
   CS = (ECS + (-CR))/2,
   \end{equation}
where we normalize $ECS$ and $-CR$ to the scale of 0 to 1.   
   For each seed $u$, we check the immediate next node $v$ and add an edge $(u,v)$ if the (ConceptNet-based) similarity between $u$ and $v$ is greater than $\bm{\theta}_{sim,ss}$\footnote{$\bm{\theta}$ denotes the set of parameters used in the model. 
}. If in this iteration, a node $v$ is not added in $G$,  
   we get the most recent predecessor $u$ for which the similarity exceed $\bm{\theta}_{sim,ss}$ and add $(u,v)$. 
  The idea is that if a word $u$ is more abstract than $v$ and if they are quite similar in terms of conceptual similarity, then word $v$ provides similar but more specific information than word $u$. Each node has a resource $\tilde{P}(u|\mathcal{I}_k)$, the confidence assigned by the Neural Network. If there is an edge from the node, some of this resource should be sent along this edge until for all  edges $(u,v) \in G$, $w_v$ becomes greater than $w_u$. We formulate the problem as a Linear Optimization problem:
{\small 
\begin{align}
  \underset{\bm{w}=(w_1,...w_{|\bm{S_k}|})}{\text{minimize}}\qquad & \sum_{(u,v) \in G} max\{w_u-w_v,0\} \nonumber \\
  \text{subject to }\qquad & \sum_{s\in \bm{S_k}} w_s= \sum_{s_k \in \bm{S_k}} \tilde{P}(\bm{s_k}|\mathcal{I}_k) \nonumber \\
  \phantom{\text{subject to }} & w_u = \tilde{P}(u|\mathcal{I}_k), u \notin G \nonumber \\
  & w_u \geq 0.5 \tilde{P}(u|\mathcal{I}_k),  \forall u \in G \nonumber
\end{align}
 } 
 
 To limit the resource a node $u$ can send, we limit the final minimum value by $0.5\text{ } \tilde{P}(u|\mathcal{I}_k)$.
   The solution provides us with the necessary weights for the set of seeds $\bm{S_k}$ in $\mathcal{I}_k$. We normalize these weights and get $\tilde{W}(\bm{S_k})$.

\section{Intermediate Results for the ``Aardvark'' Riddle}
\begin{figure}[!htpb]
     \centering
     \includegraphics[width=\textwidth]{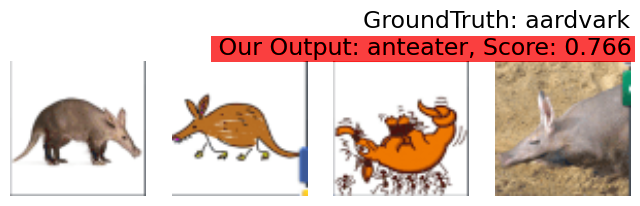}
     \vspace*{-0.1in}
     \caption{The four different Images for the ``aardvark'' riddle.}
     \label{fig:example_Riddle1}
     \vspace*{-0.1in}
\end{figure}
From the four figures in Figure \ref{fig:example_Riddle1}, we get the top 20 Clarifai detections as given in the Table \ref{tab:aardvark1}. 
\begin{table}
\begin{tabular}{|c|c|c|c|}
    \hline
    \textbf{Image1} & \textbf{Image2} & \textbf{Image3} & \textbf{Image4} \\
    \hline
 \error{monster} & \error{food} &    fun &     rock \\
 \error{jurassic} & small &   retro &   \error{nobody} \\ 
 \error{monstrous} & vector &  clip &    travel \\ 
 \error{primitive} & dinosaur & \error{halloween} & \error{water} \\ 
 lizard &  wildlife & \error{set} &     \error{sea}  \\
 paleontology & cartoon & border &  \error{aquatic}  \\
 vertebrate & nature &  messy &   outdoors  \\
 dinosaur & \error{evolution} & ink &     \error{sand}  \\
 creature & reptile & design &  \error{beach}  \\
 wildlife & outline & ornate &  bird  \\
 nature &  cute &    decoration & wildlife  \\
 \error{evolution} & sketch &  \error{ornament} & \error{biology}  \\
 reptile & painting & vector &  \error{zoology}  \\
 \error{wild} &    silhouette & \error{contour} & carnivora  \\
 horizontal & horizontal & cartoon & nature  \\
 illustration & art &     cute &    horizontal  \\
 animal &  illustration & silhouette & animal  \\
 side view & graphic & art &     side view  \\
 panoramic & animal &  illustration & panoramic  \\
 mammal &  panoramic & graphic & mammal  \\
    \hline
\end{tabular}
\caption{Top 20 detections from Calrifai API. The detections that are completely noisy is colored using red. It can be observed that the third image does not give any evidence of an animal present.}
\label{tab:aardvark1}
\end{table}
 
\begin{table}
\resizebox{\columnwidth}{!}{%
\begin{tabular}{|c|c|c|c|}
    \hline
    \textbf{Image1} & \textbf{Image2} & \textbf{Image3} & \textbf{Image4} \\
    \hline
dolphin &    graph\_toughness  &   decorative & bison\\
rhinoceros & cartography & graph\_toughness  &  american\_bison\\
\textbf{komodo\_dragon} & color\_paint & graph &      \textbf{marsupial}\\
african\_elephant & graph & artwork &    gibbon\\
\textbf{lizard} &     spectrograph &  spectrograph & monotreme\\
gorilla &    revue & kesho\_mawashi & moose\\
crocodile &  linear\_functional &  tapestry &   mole\\
indian\_elephant &  simulacrum &  map & \textbf{wildebeest}\\
\textbf{wildebeest} & pen\_and\_ink & arabesque &  \textbf{echidna}\\
elephant &   luck\_of\_draw &  sgraffito &  turtle\\
\textbf{echidna} &    \textbf{cartoon} &     linear\_functional  &  mule\_deer\\
chimaera &   camera\_lucida & hamiltonian\_graph  &  mongoose\\
chimpanzee & explode\_view &  emblazon &   tamarin\\
liger &      micrographics & pretty\_as\_picture  &  chimpanzee\\
\textbf{gecko} &      hamiltonian\_graph &   art\_deco &   wolverine\\
rabbit &     crowd\_art &   dazzle\_camouflage  &  prairie\_dog\\
\textbf{iguana} &     \textbf{depiction} &   ecce\_homo &  western\_gorilla\\
hippopotamus & echocardiogram & pointillist &  \textbf{anteater}\\
mountain\_goat & scenography & pyrography & okapi\\
loch\_ness\_monster & linear\_perspective & echocardiogram & skunk\\
    \hline
\end{tabular}
\caption{Top 20 detections per each image from PSL Stage I (GUR). }
\label{tab:aardvark2}
}
\end{table}
Based on the GUR approach (\textbf{GUR+All} in paper), our PSL Stage I outputs probable concepts (words or phrases) depending on the initial set of detected class-labels (\textit{seeds}). They are provided in Table \ref{tab:aardvark2}.
Note that, these are the top \textit{targets} detected from almost 0.2 million possible candidates. Observe the following:

i) the highlighted detected animals have a few visual features in common, such as \textit{four short legs, a visible tail, short height} etc. 

ii) the detections from the third image does not at all lead us to an animal and the PSL Stage I still thinks that its a cartoon of sort.
 
iii) the detections from second gets affected because of its close relation to the detections from third image and it infers that the image just depicts cartoon. 

In the final PSL Stage II however, the model figures out that there is an animal that is common to all these images. This is mainly because \textit{seeds} from the three images \textit{confidently} predict that some animal is present in the images. That is why most of the top detections correspond to animals and animals having certain characteristics in common.  

The top detections from PSL Stage II (GUR) are: \textit{monotreme, gecko, hippopotamus, pyrography, anteater, lizard, mule\_deer, chimaera, liger, iguana, komodo\_dragon, echidna, turtle, art\_deco, sgraffito, gorilla, loch\_ness\_monster, prairie\_dog}.

\textbf{BUR:}  For BUR, PSL Stage I outputs probable concepts (words or phrases) depending on the current set of \textit{seeds}. They are provided in the Table \ref{tab:aardvark3}. Observe that the individual detections are better compared to GUR\footnote{The output from the PSL Stage I for BUR, is completely independent of the other images. In essence, for each image, we are predicting all relevant concepts from a large vocabulary given a few detections from a small set of class-labels.}. 

\begin{table}
\resizebox{\columnwidth}{!}{%
\begin{tabular}{|c|c|c|c|}
    \hline
    \textbf{Image1} & \textbf{Image2} & \textbf{Image3} & \textbf{Image4} \\
    \hline
panda & \textbf{like\_paint} & hamiltonian\_graph &  giraffe\\
dolphin &  projective\_geometry &  graph\_toughness &  waterbuck\\
african\_forest\_elephant &  \textbf{diagram} &    lacquer &  sandy\_beach\\
placental\_mammal & line\_of\_sight &  figuration & moose\\
\textbf{otter} & venn\_diagram & war\_paint &  \textbf{wildebeest}\\
gorilla &   hippocratic\_face &   graph &    skunk\\
\textbf{wildebeest} &       real\_number\_line &  spectrograph &  \textbf{anteater}\\
\textbf{chimaera} & sight\_draft &  map &      \textbf{echidna}\\
african\_savannah\_elephant &  x\_axis &     arabesque &  bobcat\\
florida\_panther &  simulacrum & fall\_off\_analysis &  mule\_deer\\
liger & cartoon &    art\_collection  &  bison\\
rabbit &    diagrammatic & statue &   pygmy\_marmoset\\
\notice{aardvark} &  camera\_lucida &  delineate &  mongoose\\
\textbf{iguana} &  explode\_view & jack\_o\_lantern  &  sea\_otter\\
hippopotamus &     crowd\_art &  gussie\_up &  \textbf{squirrel\_monkey}\\
hadrosaur &        lottery &    ecce\_homo &  wolverine\\
mountain\_goat &    depiction &  pointillist &  okapi\\
panda\_bear &       conecept\_design &  art\_deco & cane\_rat\\
velociraptor &     infinity\_symbol &  pyrography & whale\\
whale & scenography &  scenography &  american\_bison\\
    \hline
\end{tabular}
\caption{Top 20 detections per each image from PSL Stage I (IUR). }
\label{tab:aardvark3}
}
\end{table}

Final output from PSL Stage II (for BUR) is comparable to that of the GUR approach. The top detections are: \textit{hadrosaur, sea\_otter, diagrammatic, panda, iguana, pyrography, mule\_deer, placental\_mammal, liger, panda\_bear, art\_deco, squirrel\_monkey, giraffe, echidna, otter, anteater, pygmy\_marmoset, hippopotamus}. 

   Here, the set of output mainly contains the concepts (words or phrases) that either represents ``animals with some similar visual characteristics to aardvark'' or it pertains to ``cartoon or art''.

\section{Detailed Accuracy Histograms For Different Variants}
\begin{figure}[!htb]
\vspace*{-0.15in}
     \centering
\subfloat{\includegraphics[width=0.32\textwidth,height=0.1\textheight]{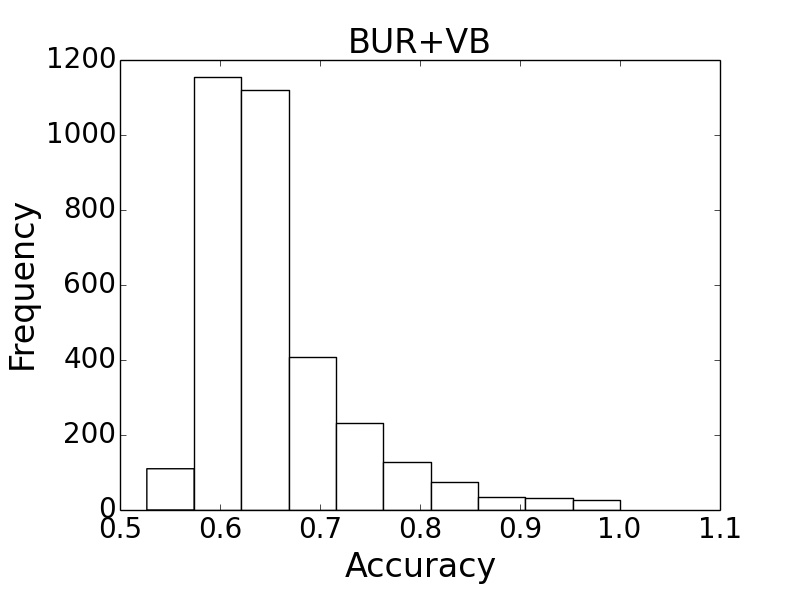}}
\hfill
\subfloat{\includegraphics[width=0.32\textwidth,height=0.1\textheight]{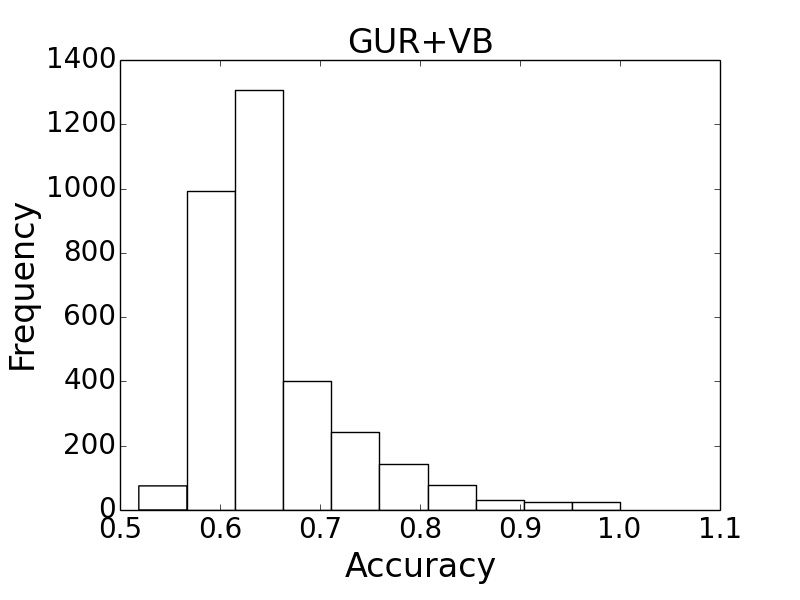} }    
\hfill
\subfloat{\includegraphics[width=0.32\textwidth,height=0.1\textheight]{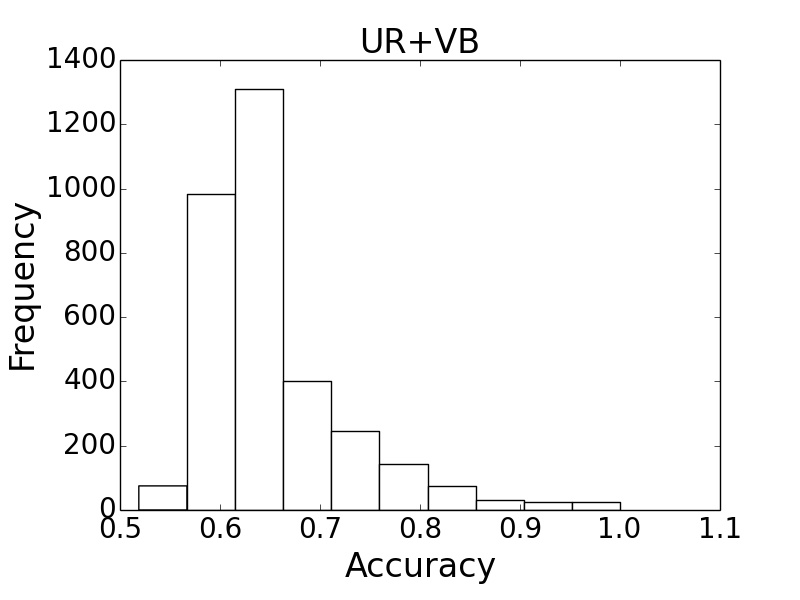} }
\hfill
\subfloat{\includegraphics[width=0.32\textwidth,height=0.1\textheight]{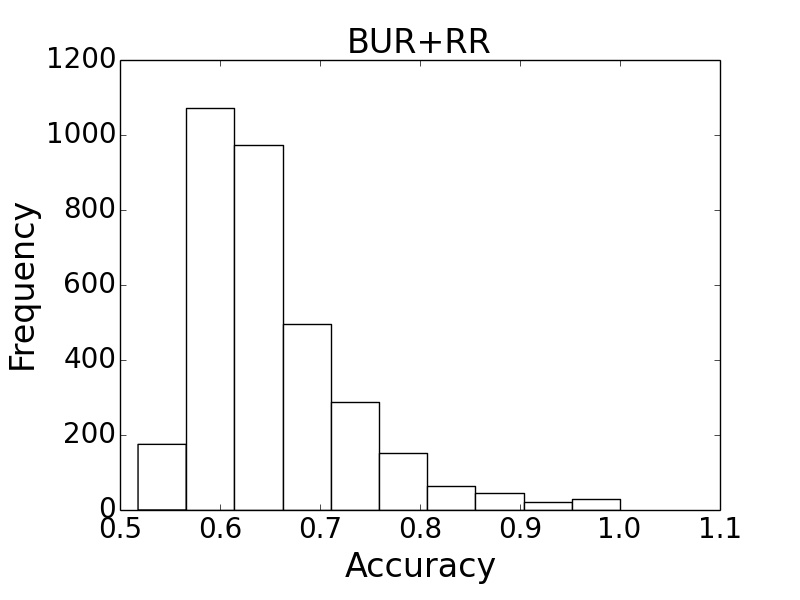}}
\hfill
\subfloat{\includegraphics[width=0.32\textwidth,height=0.1\textheight]{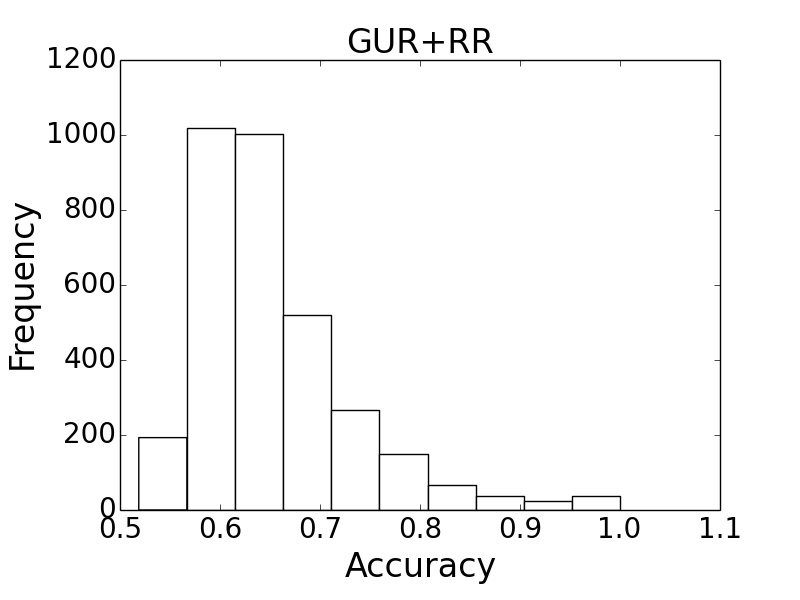} }    
\hfill
\subfloat{\includegraphics[width=0.32\textwidth,height=0.1\textheight]{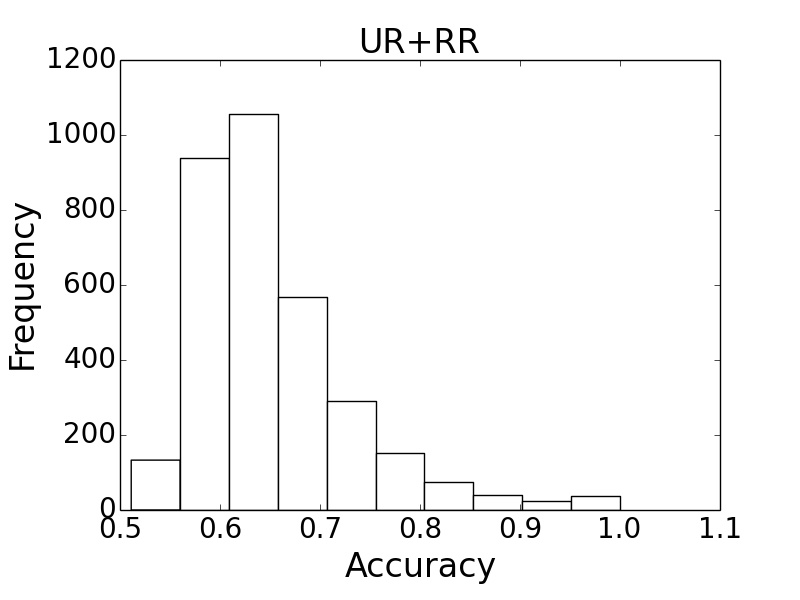} }
\hfill
\subfloat{\includegraphics[width=0.32\textwidth,height=0.1\textheight]{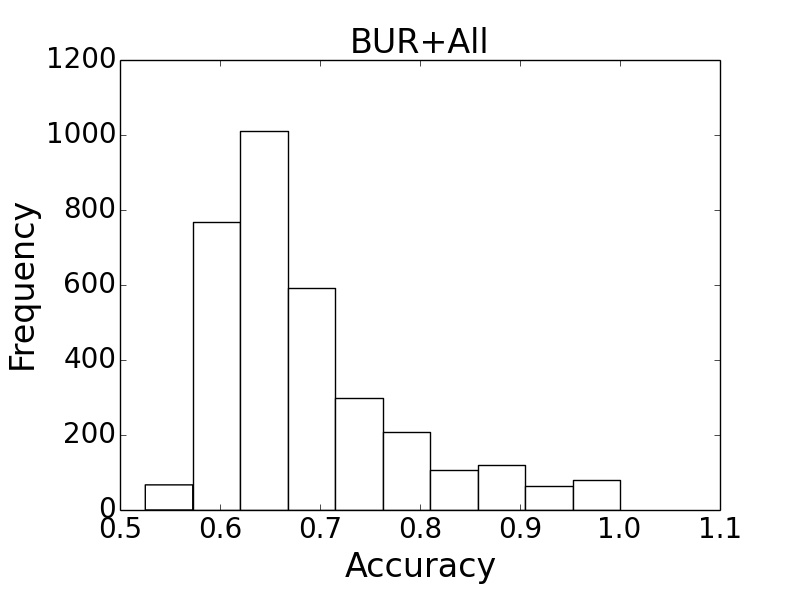}}
\hfill
\subfloat{\includegraphics[width=0.32\textwidth,height=0.1\textheight]{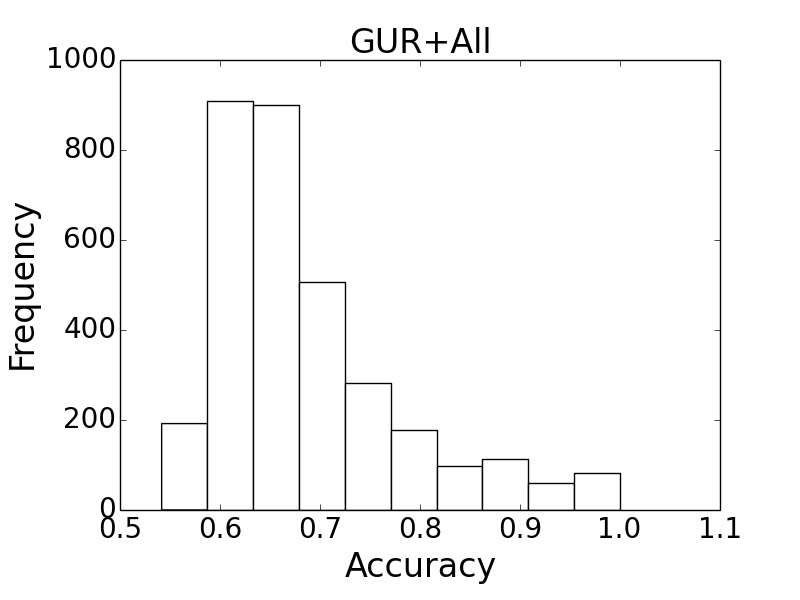} }    
\hfill
\subfloat{\includegraphics[width=0.33\textwidth,height=0.1\textheight]{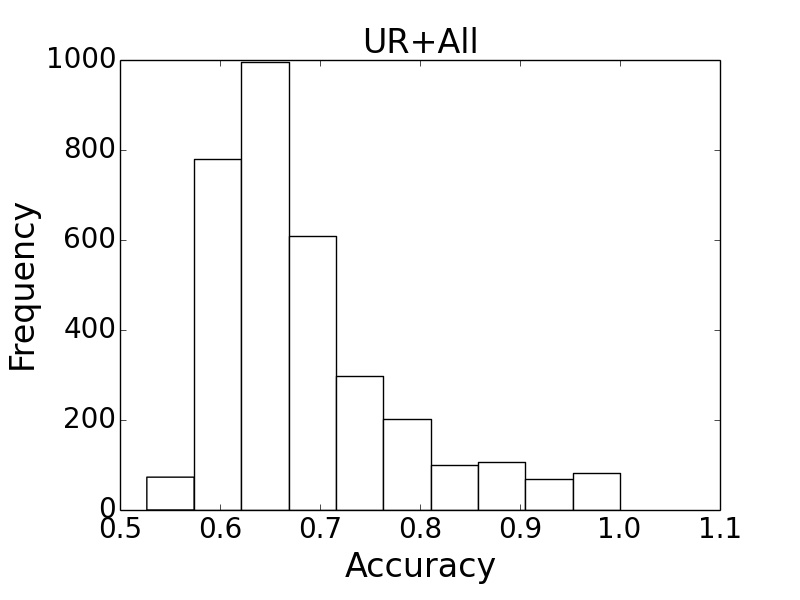} }
\caption{The accuracy histograms of the BUR, GUR and UR approaches (combined with the VB, RR and All stage variants).}
\label{fig:accuracy_histogram}
\vspace*{-0.05in}
\end{figure}

In this section, we plot the accuracy histograms for the entire dataset for all the variants (using Clarifai API) of our approach (listed in Table 2 of the paper). We also add the accuracy histograms for variants using \textbf{BUR} approach. The plots are shown in the Figure \ref{fig:accuracy_histogram}. From the plots, the shift towards greater accuracy is evident as we go along the stages of our pipeline.
  
\section{Visual Similarity: Additional Results}
Additional results for Visual Similarity are provided in Tables \ref{tab:men1} and \ref{tab:dinosaur}. 

\begin{table}[!htb]
\RawFloats
\centering
\centering
\resizebox{\columnwidth}{!}{%
\begin{tabular}{|c|c|c|}
\hline
\textbf{ConceptNet} & \textbf{Visual Similarity} & \textbf{word2vec} \\ \hline\hline
\begin{tabular}[c]{@{}c@{}}man, merby, misandrous,\\ philandry, male\_human,\\ dirty\_pig, mantyhose,\\ date\_woman,guyliner,manslut\end{tabular} & \begin{tabular}[c]{@{}c@{}}priest, uncle, guy,\\ geezer, bloke, pope,\\ bouncer, ecologist,\\ cupid, fella\end{tabular} & \begin{tabular}[c]{@{}c@{}}women, men, males,\\ mens, boys, man, female,\\ teenagers,girls,ladies\end{tabular}\\
\hline
\end{tabular}
}
\caption{Similar Words for ``Men''}
\label{tab:men1}
\end{table}
\begin{table}[!htb]
\centering
\caption{Similar Words for ``Dinosaur''}
\label{tab:dinosaur}
\resizebox{\columnwidth}{!}{%
\begin{tabular}{|c|c|c|}
\hline
\textbf{ConceptNet} & \textbf{Visual Similarity} & \textbf{word2vec}                                                                                                                                                                            \\ \hline\hline
\begin{tabular}[c]{@{}c@{}}saurischian, ornithischian,\\ protobird, elephant bird,\\ sauropsid, cassowary,\\ ibis, nightingale, ceratosaurian,\\ auk, vulture\end{tabular} & \begin{tabular}[c]{@{}c@{}}lambeosaurid, lambeosaur,\\ bird, allosauroid, therapod, stegosaur,\\ triceratops, tyrannosaurus\_rex,\\ deinonychosaur,dromaeosaur,\\ brontosaurus\end{tabular} & \begin{tabular}[c]{@{}c@{}}dinosaurs, dino, T.\_rex,\\ Tyrannosaurus\_Rex, T\_rex,\\ fossil, triceratops, dinosaur\_species,\\ tyrannosaurus,dinos,\\ Tyrannosaurus\_rex\end{tabular}
\\\hline
\end{tabular}
}
\end{table}

\section{More Positive and Negative Results}
\begin{figure*}[!htb]
\vspace*{-0.15in}
     \centering
\subfloat{\includegraphics[width=0.32\textwidth,height=0.1\textheight]{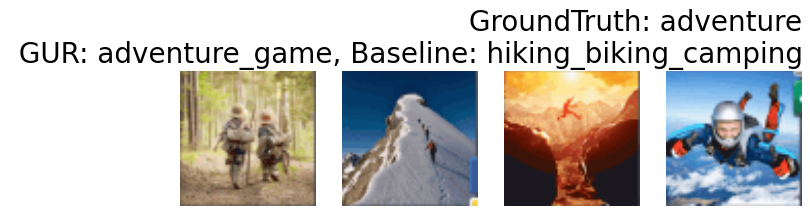} }
\hfill
\subfloat{\includegraphics[width=0.32\textwidth,height=0.1\textheight]{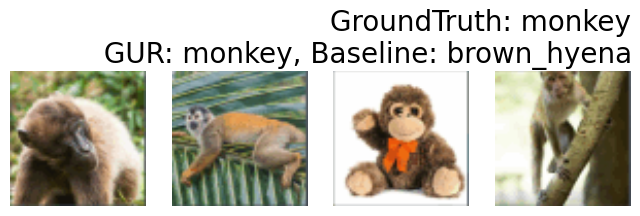}}    
\hfill
      \subfloat{\includegraphics[width=0.32\textwidth,height=0.1\textheight]{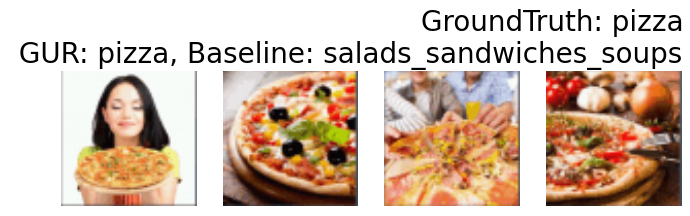} }
\hfill
\subfloat{\includegraphics[width=0.33\textwidth,height=0.1\textheight]{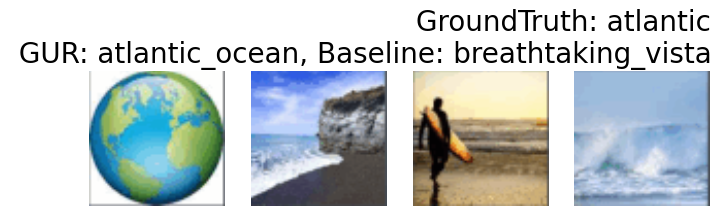}}
\hfill
\subfloat{\includegraphics[width=0.33\textwidth,height=0.1\textheight]{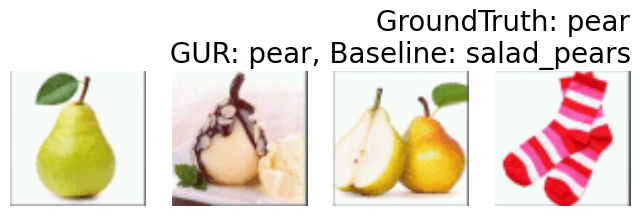}}
\hfill
\subfloat{\includegraphics[width=0.33\textwidth,height=0.1\textheight]{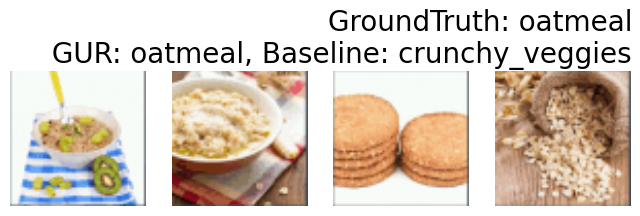}}
\hfill
\subfloat{\includegraphics[width=0.33\textwidth,height=0.1\textheight]{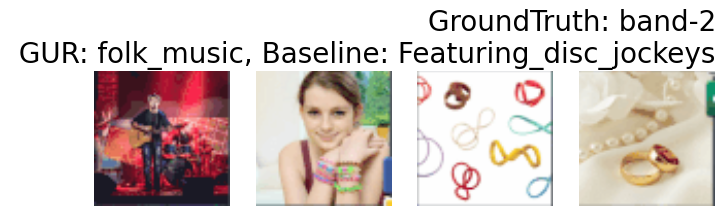}}
\hfill
\subfloat{\includegraphics[width=0.33\textwidth,height=0.1\textheight]{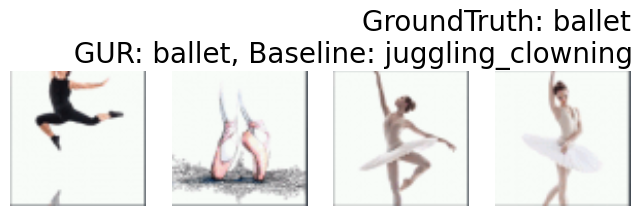}}
\hfill
\subfloat{\includegraphics[width=0.33\textwidth]{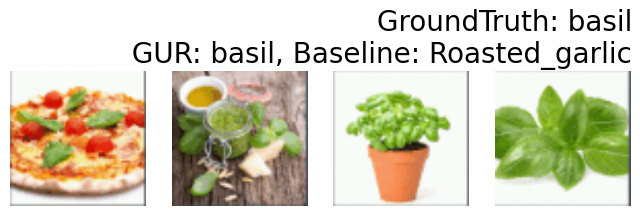}}
\hfill
\subfloat{\includegraphics[width=0.33\textwidth,height=0.1\textheight]{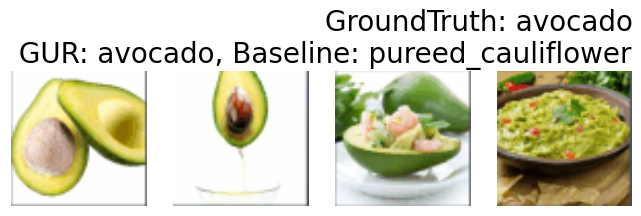}}
\hfill
\subfloat{\includegraphics[width=0.33\textwidth,height=0.1\textheight]{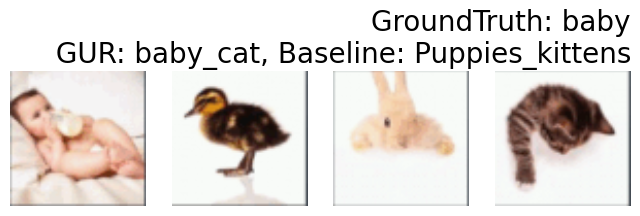}}
\hfill
\subfloat{\includegraphics[width=0.33\textwidth,height=0.1\textheight]{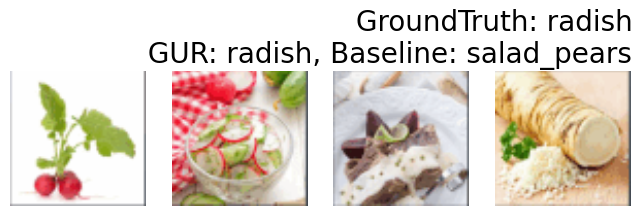}}
\hfill
\subfloat{\includegraphics[width=0.33\textwidth,height=0.1\textheight]{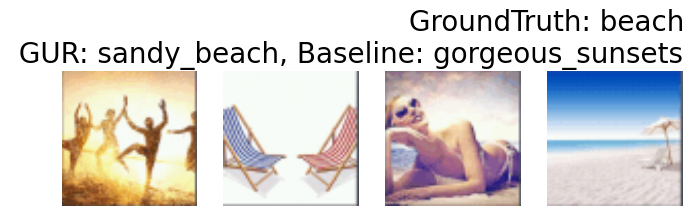}}
\hfill
\subfloat{\includegraphics[width=0.33\textwidth,height=0.1\textheight]{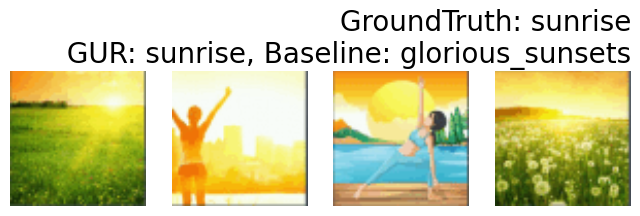}}
\hfill
\subfloat{\includegraphics[width=0.33\textwidth,height=0.1\textheight]{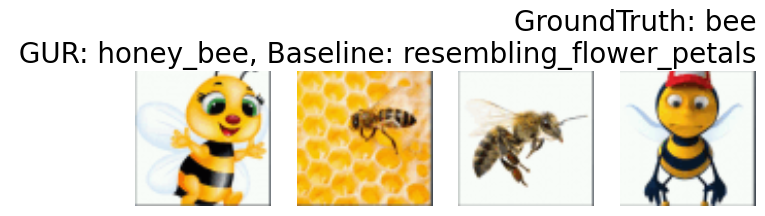}}
\hfill
\subfloat{\includegraphics[width=0.33\textwidth,height=0.1\textheight]{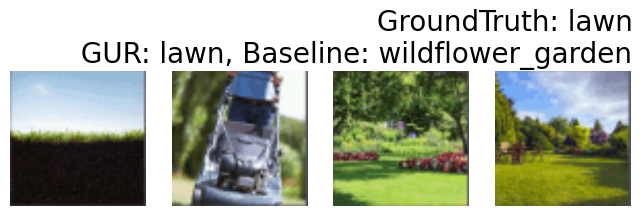}}
\hfill
\subfloat{\includegraphics[width=0.33\textwidth,height=0.1\textheight]{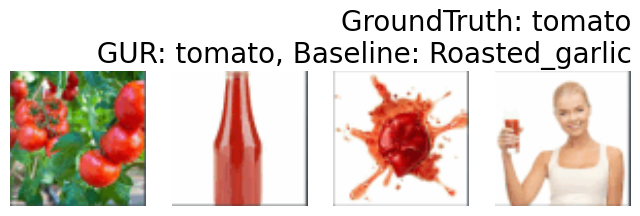}}
\hfill
\subfloat{\includegraphics[width=0.33\textwidth,height=0.1\textheight]{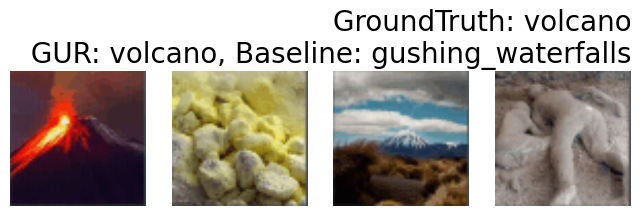}}
\hfill
\subfloat{\includegraphics[width=0.33\textwidth,height=0.1\textheight]{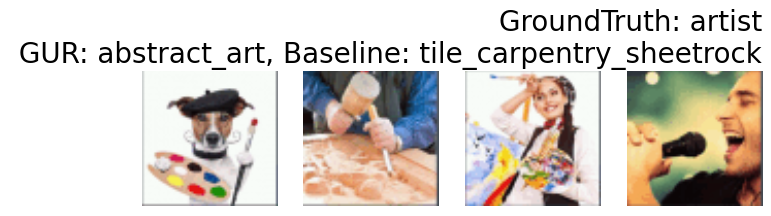}}
\hfill
\subfloat{\includegraphics[width=0.33\textwidth,height=0.1\textheight]{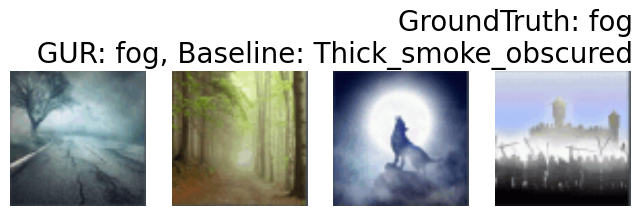}}
\hfill
\subfloat{\includegraphics[width=0.33\textwidth,height=0.1\textheight]{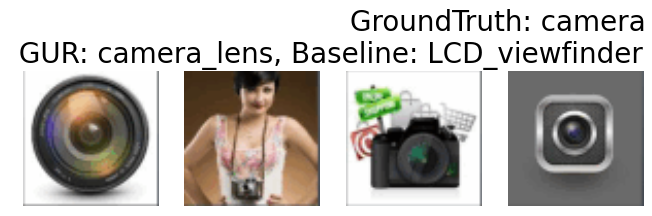}}
\hfill
      \subfloat{\includegraphics[width=0.33\textwidth,height=0.1\textheight]{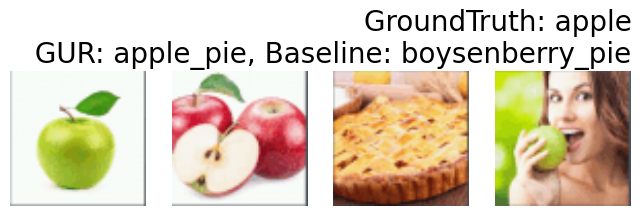} }    
\hfill
\subfloat{\includegraphics[width=0.33\textwidth,height=0.1\textheight]{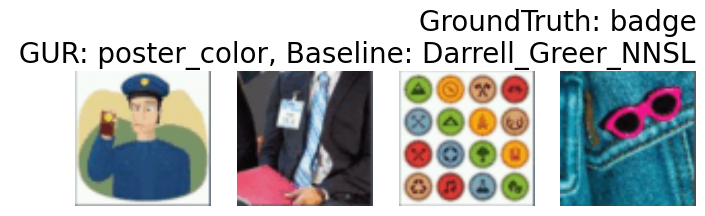}}
\hfill
\subfloat{\includegraphics[width=0.33\textwidth,height=0.1\textheight]{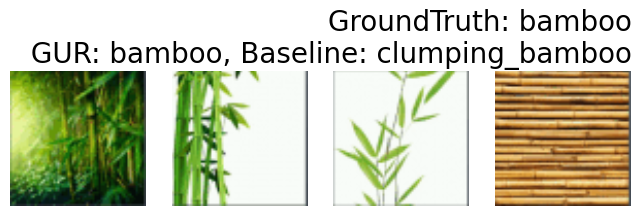}}
      \vspace*{0.1in}
     \caption{More Positive results from the ``GUR'' approach on some of the riddles. The groudtruth labels, closest label among top 10 from GUR and the Clarifai baseline are provided for all images. For more results, check \protect\url{http://bit.ly/1Rj4tFc}.}
     \label{fig:example_Riddles1}
     \vspace*{-0.2in}
\end{figure*}

\begin{figure*}[!htb]
\vspace*{-0.15in}
     \centering
\subfloat{\includegraphics[width=0.45\textwidth]{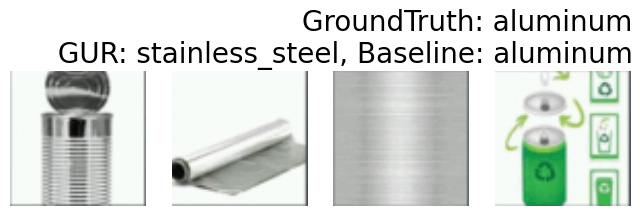}}
\hfill
\subfloat{\includegraphics[width=0.45\textwidth]{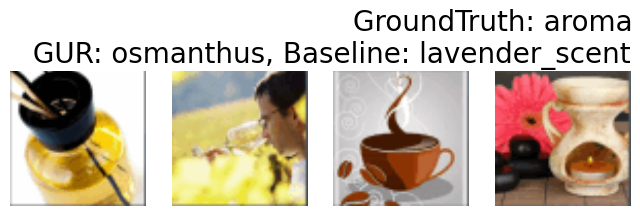} }    
\hfill
      \subfloat{\includegraphics[width=0.45\textwidth]{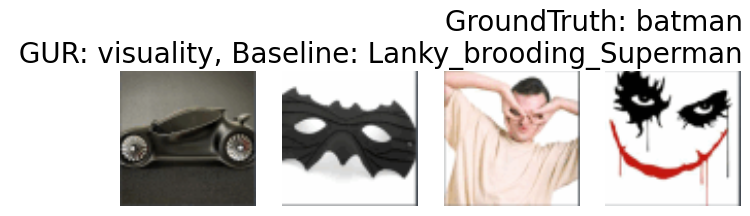} }
\hfill
      \subfloat{\includegraphics[width=0.45\textwidth]{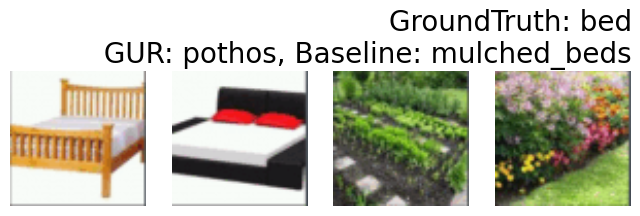} } 
\hfill
      \subfloat{\includegraphics[width=0.45\textwidth]{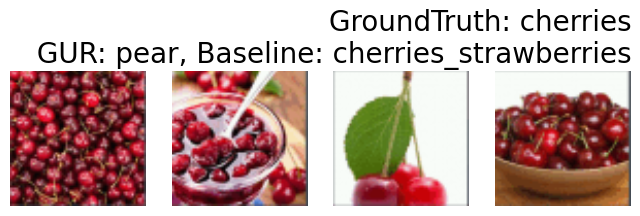} }
\hfill
      \subfloat{\includegraphics[width=0.45\textwidth]{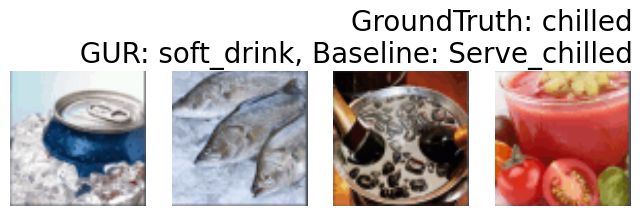} }      
\hfill
      \subfloat{\includegraphics[width=0.45\textwidth]{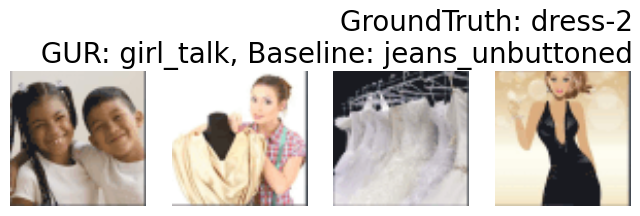} }
\hfill
      \subfloat{\includegraphics[width=0.45\textwidth]{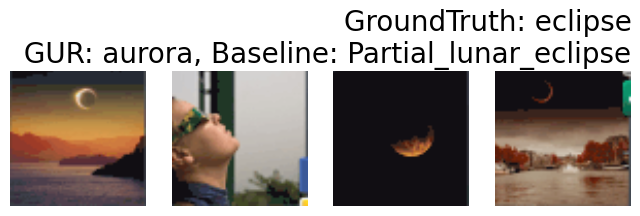} }    
\hfill
      \subfloat{\includegraphics[width=0.45\textwidth]{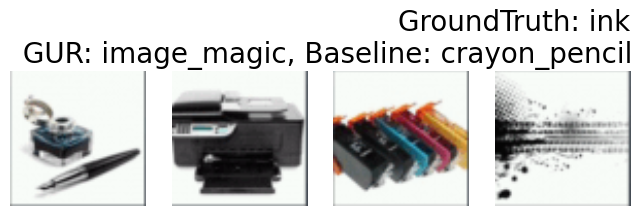} }
\hfill
      \subfloat{\includegraphics[width=0.45\textwidth]{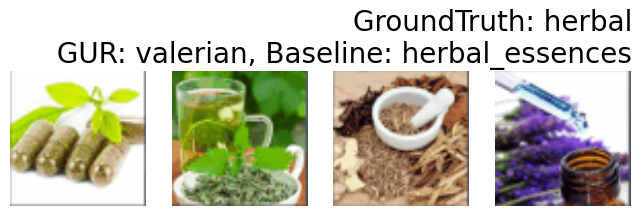} }
\hfill
      \subfloat{\includegraphics[width=0.45\textwidth]{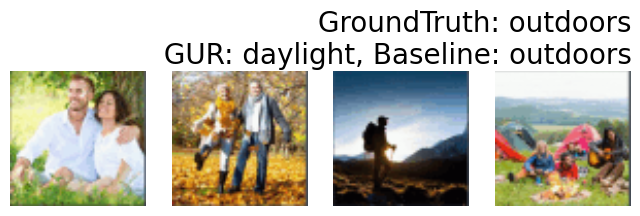} }
\hfill
      \subfloat{\includegraphics[width=0.45\textwidth]{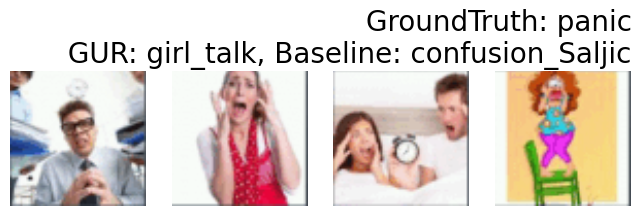} }        
      \vspace*{0.1in}
     \caption{Some Negative results from the ``GUR'' approach on some of the riddles. The groudtruth labels, closest label among top 10 from GUR and the Clarifai baseline are provided for all images. For more results, check \protect\url{http://bit.ly/1Rj4tFc}.}
     \label{fig:example_Riddles2}
     \vspace*{-0.2in}
\end{figure*}

We provide positive and Negative results in Figures \ref{fig:example_Riddles1} and \ref{fig:example_Riddles2} of the "GUR+All" variant of the pipeline. We obtain better results with Clarifai detections rather than Residual Network detections. Based on our observations, one of the key property of the ResidualNetwork confidence score distribution is that there are few detections (1-3) which are given the strongest confidence scores and the other detections have very negligible confidence scores.  These top detections are often quite noisy.

 For example, for the aardvark image 1, the ResidualNetwork detections are: \textit{\textbf{triceratops}, wallaby, armadillo, hog, fox squirrel, wild boar, kit fox, grey fox, Indian elephant, red fox, mongoose, Egyptian cat, wombat, tusker, mink, Arctic fox, toy terrier, dugong, lion}. Only the first detection has $0.84$ score and the rest of the scores are very negligible. For the second, third and fourth images, the top detections are respectively:
\begin{enumerate}
\item {\bf \color{red}pick} ($0.236$), ocarina ($0.114$), maraca ($0.091$), chain saw ($0.06$), whistle ($0.03$), {\bf \color{red}can opener} ($0.03$), \textbf{triceratops} ($0.02$), muzzle, spatula, loupe, hatchet, letter opener, thresher, rock beauty, electric ray, tick, gong, Windsor tie, cleaver, electric guitar

\item {\bf \color{red}jersey} ($0.137$), {\bf \color{red}fire screen} ($0.129$),  {\bf \color{red}sweatshirt} ($0.037$), pick ($0.035$),  \textbf{comic book} ($0.030$), book jacket ($0.029$), plate rack, throne, wall clock, face powder, binder, hair slide,velvet,puck, redbone.

\item {\bf \color{red}hog}  ($0.48$), wallaby  ($0.19$), wild boar  ($0.10$), Mexican hairless  ($0.045$), gazelle  ($0.023$), wombat ($0.017$), dhole ($0.016$), hyena ($0.015$), \textbf{armadillo} ($0.009$), ibex, hartebeest, water buffalo, bighorn, kit fox, \textbf{mongoose}, hare, wood rabbit,  warthog, mink, polecat. 
\end{enumerate}
  These predictions show that for the first and fourth image, there are some animals detected with some distant visual similarities. The second and third image has almost no animal mentions. This also shows some very confident detections (such as \textbf{triceratops} for the first image) is quite noisy.

In many cases, due to these high-confidence noisy detections, the PSL-based inference system gets biased towards them. Compared to that, Clarifiai detections provide quite a few (abstract but) correct detections about different aspects of the image (for example, for 2nd Image, predicts labels related to ``cartoon/art'' and ``animal'' both). This seems to be one of the reasons, for which the current framework provide better results for Clarifai Detections. Using Residual Network, the final output from the GUR system for the ``aardvark'' riddle is: \textit{antelope, prairie\_dog,  volcano\_rabbit, marsupial\_lion, peccary, raccoon,
pouch\_mammal, rabbit, \textbf{otter}, monotreme, jackrabbit, hippopotamus, moose, \textbf{tapir}, \textbf{echidna}, gorilla}.

\end{document}